%% file: main.tex
\pdfoutput=1

\documentclass[10pt,journal,cspaper,compsoc]{IEEEtran}
\usepackage[utf8]{inputenc} % allow utf-8 input
\usepackage[T1]{fontenc}    % use 8-bit T1 fonts
\usepackage{url}            % simple URL typesetting
\usepackage{booktabs}       % professional-quality tables
\usepackage{amsfonts}       % blackboard math symbols
\usepackage{nicefrac}       % compact symbols for 1/2, etc.
\usepackage{microtype}      % microtypography
\usepackage{amssymb}
\usepackage{amsmath}
\usepackage{amsthm,subfigure}
\usepackage{graphicx}
\usepackage{algpseudocode}
\usepackage[linesnumbered,ruled]{algorithm2e}
\usepackage{xfrac}
\usepackage{color}
\usepackage{wrapfig,amsfonts}
\usepackage{paralist}
\usepackage{mathtools}
\usepackage{comment}
\usepackage{hyperref}
\usepackage[svgnames]{xcolor}
\usepackage[customcolors,shade]{hf-tikz}
\newcommand{\subparagraph}{}
\usepackage{empheq,multicol,titlesec}
\usepackage{enumitem}

\usepackage[most]{tcolorbox}
\newtcbox{\mymath}[1][]{%
    nobeforeafter, math upper, tcbox raise base,
    enhanced, colframe=blue!30!black,
    colback=gray!30, boxrule=1pt,
    #1}

\usepackage{multirow}
\usepackage{colortbl, array}
\usepackage{hhline}

\definecolor{rulecolor}{RGB}{0,71,171}
\definecolor{tableheadcolor}{RGB}{204,229,255}

\makeatletter
\newsavebox{\mybox}\newsavebox{\mysim}
\newcommand{\distras}[1]{%
  \savebox{\mybox}{\hbox{\kern3pt$\scriptstyle#1$\kern3pt}}%
  \savebox{\mysim}{\hbox{$\sim$}}%
  \mathbin{\overset{#1}{\kern\z@\resizebox{\wd\mybox}{\ht\mysim}{$\sim$}}}%
}

\newtheorem{hyp}{Hypothesis}

\usepackage{xcolor,colortbl}

\definecolor{Gray}{gray}{0.85}
\definecolor{LightCyan2}{rgb}{0.94,0.81,0.81}
\definecolor{LightCyan}{rgb}{0.94,0.39,0.39}

\newcolumntype{a}{>{\columncolor{Gray}}c}
\newcolumntype{b}{>{\columncolor{white}}c}

\ifCLASSINFOpdf
\else
\fi

\begin{document}
\title{Spatial Transformer for 3D Point Clouds}

\author{
\setlength{\tabcolsep}{20pt}
\begin{tabular}{@{}ccc@{}}
Jiayun Wang &
 Rudrasis Chakraborty &
     Stella X. Yu
\end{tabular}     
\IEEEcompsocitemizethanks{\IEEEcompsocthanksitem The authors are with UC Berkeley / ICSI, 2150 Shattuck Ave, Berkeley, CA, 94704. E-mail: \{peterwg, rudra, stellayu\}@berkeley.edu. Corresponding author: Stella X. Yu. \IEEEcompsocthanksitem Our code us publicly available at \url{http://pwang.pw/spn.html}.}% <-this % stops an unwanted space
}


\markboth{}%
{Wang \MakeLowercase{\textit{et al.}}: Spatial Transformer for 3D Point Clouds}



\IEEEtitleabstractindextext{%
\begin{abstract}
  \input{sections/0abstract.tex}
\end{abstract}

}

\maketitle

\IEEEdisplaynontitleabstractindextext

% \fi
\IEEEpeerreviewmaketitle

\input{sections/1intro.tex}

\input{sections/2related.tex}

\input{sections/3model.tex}

\input{sections/4results.tex}

\input{sections/5concl.tex}




%\bibliographystyle{IEEEtran}
\bibliographystyle{ieeetran}
\bibliography{egbib}
\input{biography.tex}

\end{document}

%% file: sections/0abstract.tex
Deep neural networks are widely used for understanding 3D point clouds. At each \textit{point convolution layer}, features are computed from local neighbourhoods of 3D points and combined for subsequent processing in order to extract semantic information.
Existing methods adopt the same  individual point neighborhoods throughout the network layers, defined by the same metric on the fixed input point coordinates. 
This common practice is easy to implement but not necessarily optimal. 
Ideally, local neighborhoods should be different at different layers, as more latent information is extracted at deeper layers.
We propose a novel end-to-end approach to learn different non-rigid transformations of the input point cloud so that optimal local neighborhoods can be adopted at each layer. 
We propose both linear (affine) and non-linear (projective and deformable) spatial transformers for 3D point clouds. With spatial transformers on the ShapeNet part segmentation dataset, the network achieves higher accuracy for all categories, with 8\% gain on earphones and rockets in particular. Our method also
outperforms the state-of-the-art on  other point cloud tasks such as classification, detection, and semantic segmentation. Visualizations show that spatial transformers can learn features more efficiently by dynamically altering local neighborhoods according to the geometry and semantics of 3D shapes in spite of their within-category variations. 

%% file: sections/1intro.tex
\IEEEraisesectionheading{\section{Introduction}\label{sec:introduction}}
\begin{figure*}[htb]
  \centering
  \includegraphics[width=.81\textwidth]{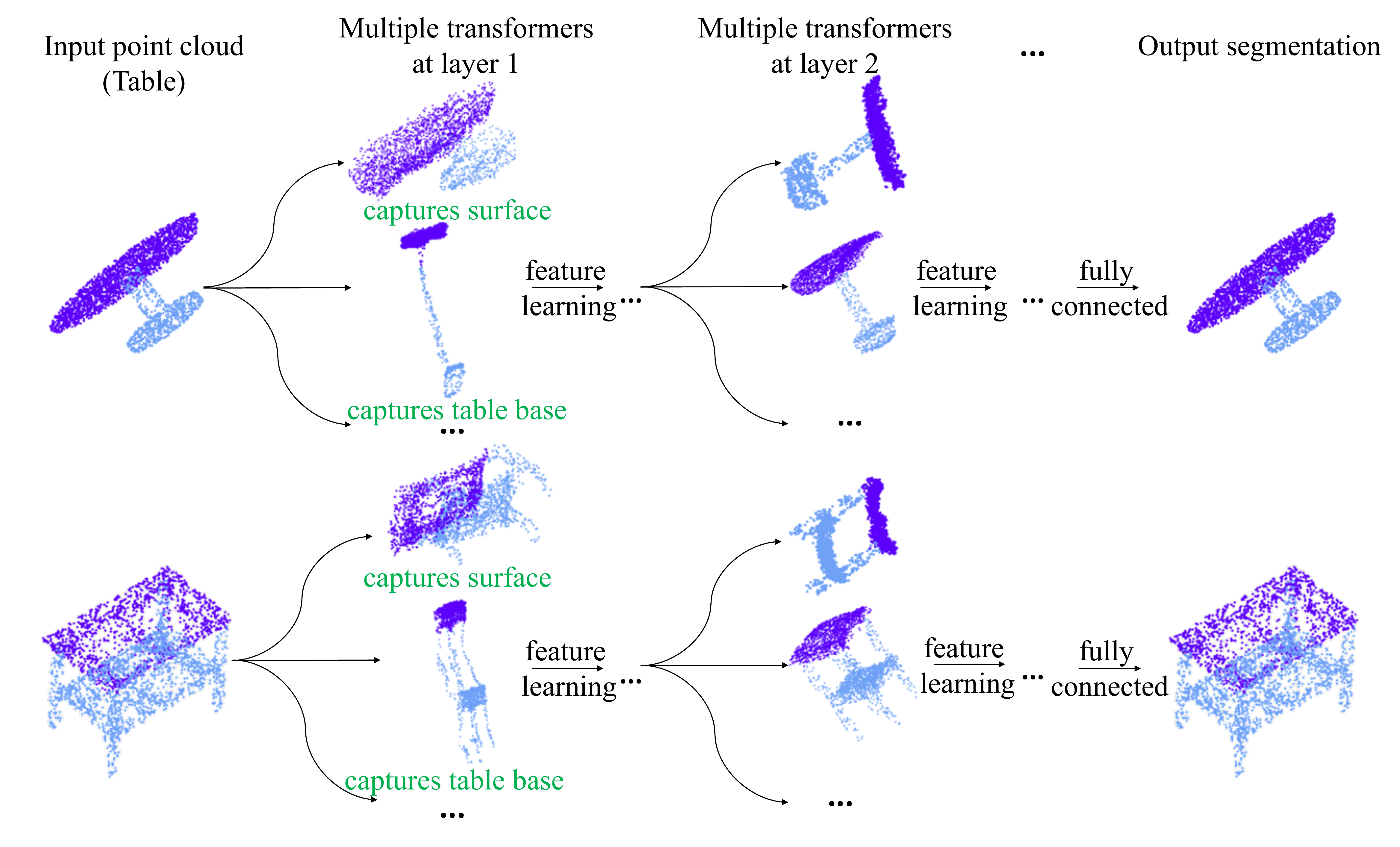}
  \caption{
  Spatial transformers learn several \textit{global transformations} at each layer to obtain different local point neighborhoods.
  We show transformed point clouds at different layers learned by spatial transformers for different instances of a category (e.g. tables).
  Compared with previous works adopting fixed local neighborhoods, dynamic point neighborhoods make the network more powerful in learning semantics from point clouds.
  For example, corresponding geometric transformations capture similar semantic information even high intra-class spatial variations exist.
  The second transformation at layer 1 deforms different tables to be more semantically similar, and makes parsing the part of table base easier.
  Furthermore, the proposed transformer is a stand-alone module and can be easily added to existing point cloud processing networks.}
  \vspace{-1em}
  \label{fig:overview}
\end{figure*}

\begin{figure}[htb]
  \centering
  \includegraphics[width=0.5\textwidth]{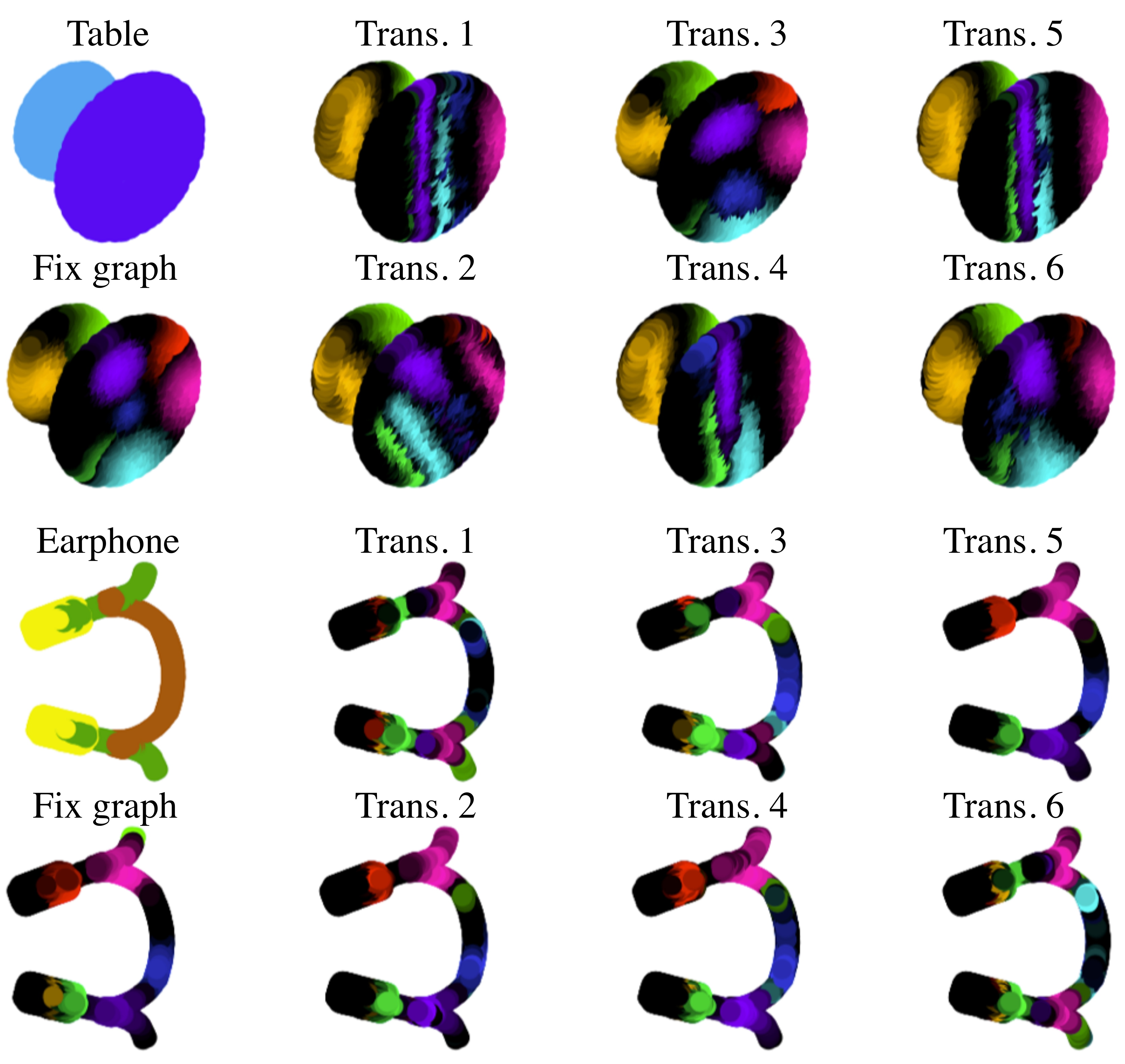}
    \caption{At each layer, we apply multiple spatial transformers to deform the input point cloud for learning different neighborhoods.
    We show local neighborhoods of  input point cloud examples with and without deformable  transformers. Different colors indicate different neighborhoods, and intensities indicate distances to the central point. 
    The dynamic neighborhood enhance the network capacity to learn from objects with large spatial variations. Rotating \href{https://drive.google.com/file/d/1261A3Pgnx8FWnZdiKLm6qReQev_j_aSK/view?usp=sharing}{table} and \href{https://drive.google.com/file/d/11iaI_rKzueMRdlPCpFFBaDBZeLABFMwV/view}{earphone} for better visualizations.
    }
    \vspace{-1em}
  \label{fig:NN2}
\end{figure}

\IEEEPARstart{3D} computer vision has been on the rise with more advanced 3D sensors and computational algorithms. Depth cameras and LiDAR sensors output 3D point clouds, which become key components in several 3D computer vision tasks including but not limited to virtual / augmented reality \cite{lin2018novel, rambach2016learning}, 3D scene understanding \cite{tulsiani2018factoring, vasu2018occlusion, dai2018scancomplete}, and autonomous driving \cite{chen2018lidar, li2018stereo, wu2018squeezeseg}.

On the algorithmic side, convolutional neural networks (CNNs) have achieved great success in many computer vision tasks \cite{deng2009imagenet, krizhevsky2012imagenet}.  However, the concept of convolution cannot be directly applied to a point cloud, as 3D points are not pixels on a regular grid with regular neighbourhoods.  One line of approaches is to convert the 3D point cloud into a representation where CNNs are readily applicable, e.g.,
a regular voxel representation \cite{wu20153d, riegler2017octnet, wang2017cnn} or 2D view projections \cite{su2015multi, qi2016volumetric,kalogerakis20173d, zhou2018learning}. 

Another line of approaches is to develop network architectures that can directly process point clouds \cite{pointnet, pointnet++, rethage2018fully, gadelha2018multiresolution}.  Analogous to convolution on 2D pixels,  convolution on 3D points  needs to first identify \textit{local neighborhoods} around individual input points.  This step is achieved by computing the so-called {\it point affinity matrix}, i.e., the adjacency matrix of a dense graph constructed from the point cloud.  These neighborhoods are then used for extracting features with point-wise convolutions.  By stacking basic point convolution layers, a neural network can extract information from the point cloud with an increasing level of abstraction.  

However, unlike images where 2D pixels are laid out on a regular grid with simple and well-defined local neighborhoods, local neighborhoods of 3D points are ill-defined and subject to various geometric transformations of 3D shapes. Most methods \cite{pointnet, pointnet++,DGCNN, li2018pointcnn} define local neighborhoods as nearest neighbors  in the Euclidean space of the input 3D point coordinates. %$\ell_2$ metric. 

This common practice of defining a
nearest neighbor graph according to the Euclidean distances on the fixed input 3D point coordinates
may be simple but not optimal.  
First, such distances may not be able to efficiently encode semantics of 3D shapes, e.g., semantically or topologically far points might be spatially close in terms of the Euclidean distances.
Secondly, fixed neighborhoods throughout the network may reduce the model's learning capacity as different layers capture information at different levels of abstraction, e.g., objects have a natural hierarchy and in order to segment out their parts, it would be more efficient to provide different layers the ability to parse them at different spatial scales.

We propose to address these fixed point neighbourhood restrictions by dynamically learning local neighborhoods and transforming the input point cloud at different layers.
We use a parametric model that takes both point coordinates and learned features as inputs to learn the point affinity matrix.
At different layers of the network, we learn several different transformations (dubbed as \textit{spatial transformers} or \textit{transformers} hereafter, Fig.\ref{fig:overview}) and corresponding point local neighborhoods (Fig.\ref{fig:NN2}). Spatial transformers allow the network to adaptively learn point features covering different spatial extensions at each depth layer.

To spatially transform a point cloud, we learn a function, $\Phi$, that generates transformed point coordinates from the input point coordinates and feature maps.  However,
it is nontrivial to learn $\Phi$ without smoothness constraints.  
Since any isometric (e.g. rigid) transformation  cannot change the distance metric, 
we consider non-rigid transformations, both linear and non-linear families.
That is, our \textit{spatial transformers} are parameterized functions conditioned on the input point coordinates $P$ and feature map $F$; they are subsequently used to transform the point coordinates, resulting in a new point affinity matrix for obtaining dynamic local neighborhoods.

We consider three families of spatial transformers.
\begin{inparaenum}[\bf 1)]
\item Affine transformation $P \mapsto AP$, where $A$ is an affine matrix. 
\item Projective  transformer $\widetilde{P} \mapsto B\widetilde{P}$, where $\widetilde{P}$ is 3D points expressed in homogeneous coordinates. 
\item Deformable transformer $P \mapsto CP + DF$, where $C, D$ are respective transformation matrices of point coordinates and features $F$. 
The transformation depends on both the input point coordinates and the features the points assume.
\end{inparaenum}

Our work makes the following contributions. \begin{itemize} 
\item We propose to learn linear (affine) and non-linear (projective, deformable)  spatial transformers for new point affinity matrices and thus dynamic local neighborhoods throughout the neural network.
\item We demonstrate that our spatial transformers can be easily added to existing point cloud networks for a variety of tasks: classification, detection, and segmentation. 
\item We apply spatial transformers to various point cloud processing networks, with point-based and sampling-based metrics for point neighborhoods, and observe performance gains of dynamic graphs over fixed graphs. 
\end{itemize}

%% file: sections/2related.tex
\section{Related Work}
We discuss related works that motivate the necessity of our proposed spatial transformers. 

\noindent \textbf{View-based methods} project 3D shapes to 2D planes and use images from multiple views as representations. Taking advantages of the power of CNNs in image processing \cite{feng2018gvcnn, han2019seqviews2seqlabels, qi2016volumetric, su2015multi}, view-based methods achieve reasonable 3D processing performance.
However, certain information about 3D structures gets lost when 3D points are projected to 2D image planes; occluded surfaces and density variations are thus often troublesome for these methods.

\noindent \textbf{Voxel-based methods} 
represent 3D shapes as volumetric data on a regular 3D grid, and proceed with 3D convolution \cite{wu20153d, maturana2015voxnet, tatarchenko2017octree}.  Their caveates are quantization artifacts, inefficient usage of 3D voxels, and low spatial resolutions due to a large memory requirement.  In addition,
3D convolutions are not biased towards surface property extraction and thus cannot capture geometrical and semantic information efficiently. 
Recent works that apply different partition strategies \cite{klokov2017escape, riegler2017octnet, tatarchenko2017octree, wang2017cnn} relieve these issues but depend heavily on bounding volume subdivision instead of local geometric shapes.  In contract, our method works directly on the 3D point cloud, minimizing geometric information loss and maximizing processing efficiency.

\noindent \textbf{Point cloud processing methods} take a point cloud as the input and extract semantic information by point convolutions.  PointNet \cite{pointnet} directly learns the embedding of every 3D point in isolation and gather that information by pooling point features later on.  Although it achieves good performance at the time, PointNet does not learn any 3D local shape information since each local neighborhood contains only one point.  PointNet++ \cite{pointnet++} addresses this caveate by adopting a hierarchical application of isolated 3D point feature learning to multiple subsets of a point cloud.  Many other works also explore different strategies for leveraging local structure learning from point clouds \cite{DGCNN, li2018pointcnn}.  Instead of finding neighbors of each point, SplatNet \cite{su2018splatnet} encodes local structures from the sampling perspective: it groups points based on  permutohedral lattices \cite{adams2010fast}, and then applies bilateral convolution \cite{bilaterCNN} for feature learning.  Super-point graphs \cite{landrieu2018large} partition a point cloud into super-points and learn the 3D point geometric organization. 
Many works focus on designing novel point convolutions given 3D point local neighborhoods \cite{pointnet++, su2018splatnet, li2018pointcnn}, ignoring how the local neighborhoods should be formed.

Unlike pixels in a 2D image, points in a 3D point cloud are un-ordered, with irregular and heterogeneous neighborhoods; regular convolution operations thus cannot be applied. Many works \cite{DGCNN, li2018pointcnn, wang2018deep, tatarchenko2018tangent, zhao2019pointweb, li2018so} aim to design point convolution operations that resemble regular 2D convolutions.  Fixed input point coordinates are used to define local neighborhoods for the point convolution, resulting in the same local neighbourhoods at different layers that limit the model's processing power.
In contrast, our work uses spatial transformers at each layer to learn dynamic local neighborhoods in a more adaptive, flexible, and efficient way.

\noindent {\bf Spatial Transformations.} The idea of enabling spatial transformation in neural networks has been explored for 2D image understanding \cite{jaderberg2015spatial}.  It is natural to extend the idea to 3D point clouds.
PointNet \cite{pointnet} adopts a rigid transformation module on the input point cloud to factor out the object pose and improve classification accuracy. 
KPConv \cite{thomas2019kpconv} applies \textit{local deformation} in the neighborhood of point convolution to enhance its learning capacity. 
In contrast, our work learns several different global transformations to apply on the input point cloud at each layer for dynamic neighborhoods.

%% file: sections/3model.tex
\section{Methods}

We first briefly review different geometric transformation methods and their influence on the affinity matrix of point cloud data, then describe the design of our three \textit{spatial transformers}, namely, \begin{inparaenum}[\bfseries (a)] \item affine, \item projective and \item deformable. \end{inparaenum}
We apply the \textit{spatial transformer block}, consisting of multiple spatial transformers, to each layer of a network for altering local neighborhoods for better point feature learning. 
We conclude the section by introducing how the transformers can be added to existing point cloud processing networks and the relevance to other works.

\subsection{Geometric Transformations}
We propose to learn transformations on the input point cloud to \textit{deform} its geometric shape, and alter local neighborhoods with new point affinity matrices. 
The hypothesis behind the usage of geometric transformation is as follows:

\begin{hyp}
Let $P=\left\{\mathbf{p}_i\right\}$ be the input point cloud and let $\mathcal{N}_i$ be the local neighborhood around $\mathbf{p}_i \in \mathbf{R}^3$ from which we extract local features. Let $\mathcal{N} = \left\{\mathcal{N}_i\right\}$ be the set of local neighborhoods. Assume $\widetilde{\mathcal{N}} = \left\{ \widetilde{\mathcal{N}}_i \right\}$ be the \textit{optimal neighborhood} for learning local features, then $\exists (\text{smooth})\:\Phi: \mathcal{N}_i \rightarrow \widetilde{\mathcal{N}}_i$ for all $\mathbf{p}_i$. 
\end{hyp}
 Essentially we are going to use different types of geometric transformations to approximate $\Phi$. The new learned affinity matrix will dynamically alter local neighborhoods to allow better feature learning.

  \begin{figure}
    \centering
    \includegraphics[width=0.48\textwidth]{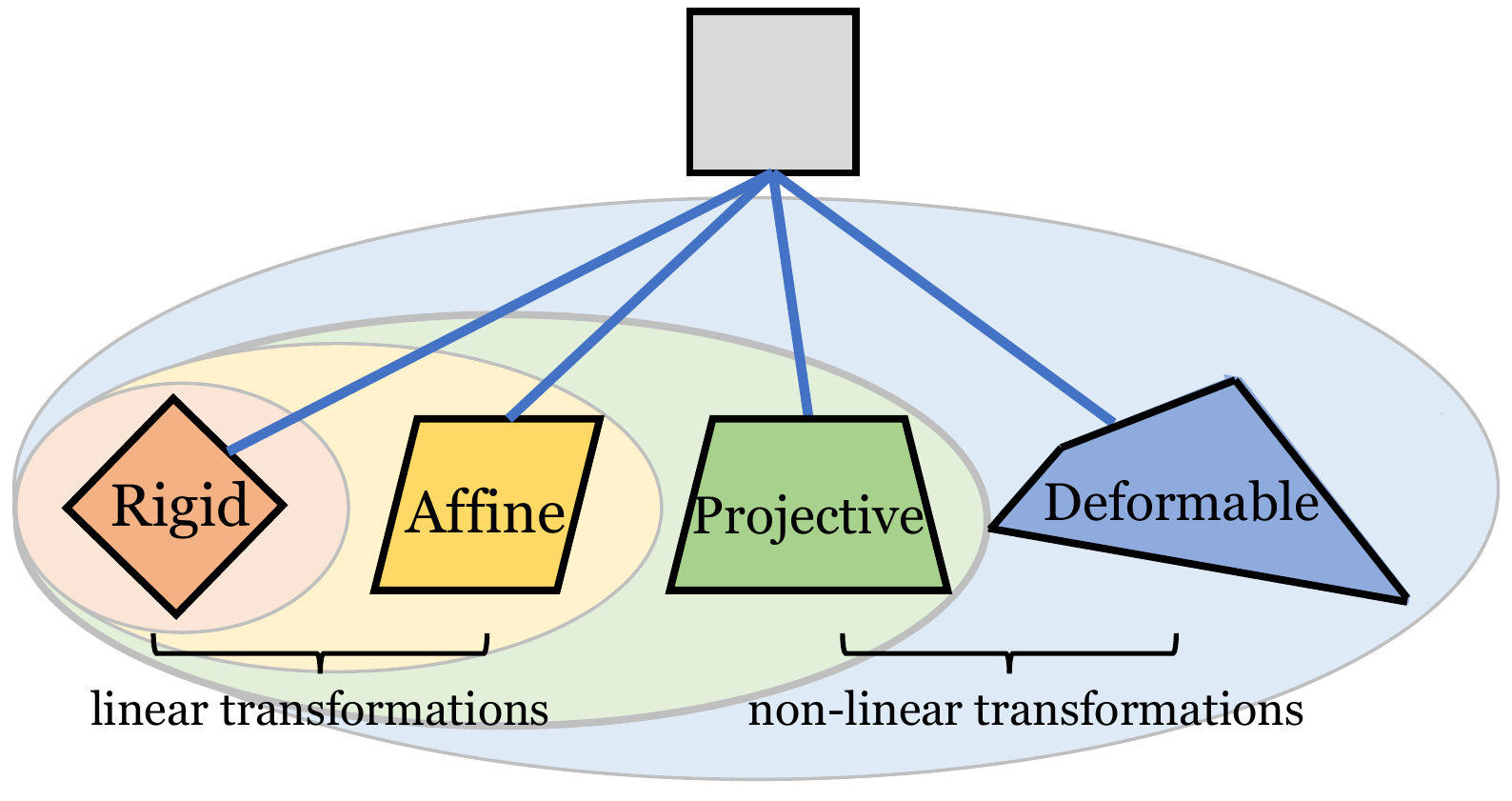}%
   \caption{ Geometric transformations. We illustrate how a grey square transforms after rigid, affine, projective and deformable transformations.}
   \label{fig:transformation}
   \vspace{-1em}
  \end{figure}

Illustrated in Fig.\ref{fig:transformation}, transformations can be categorized into rigid and non-rigid transformations, and the latter can be further categorized into linear and non-linear transformations. We now discuss different spatial transformations.

\noindent {\bf Rigid Transformations.} The group of rigid transformations consist of translations and rotations. However, rigid transformations are isometric (in $\ell_2$ distance) and therefore preserves the point affinity matrix. Thus, local neighborhoods are \textit{invariant} to rigid transformations in terms of $k$-NN graphs. Hence, we do not consider rigid transformations.

\noindent {\bf Affine Transformations.} Affine transformations belong to non-rigid linear transformations. Consider a 3D point cloud  $P  = \{\mathbf{p}_i\}_{i=1}^N\subset  \mathbf{R}^{3}$ consisting of $N$ three-dimensional vectors $\mathbf{p}_i \in \mathbf{R}^{3}$. Then, an affine transformation can be parameterized by an invertible matrix $A \in \mathbf{R}^{3 \times 3} $ and a translation vector $\mathbf{b} \in \mathbf{R}^{3} $. Given $A, \mathbf{b}$, the affine transformed coordinates  $\mathbf{p}_i$ can be written as $\mathbf{p}_i \mapsto A\mathbf{p}_i + \mathbf{b}$. Note that translation $\mathbf{b}$ does not change the point affinity matrix and point neighborhoods. Recall that an affine transformation preserves collinearity, parallelism, and convexity.

\noindent {\bf Projective Transformations.}  Projective transformations are  non-rigid non-linear transformations.  We first map the 3D point cloud $P$ to the homogeneous space and get $\widetilde{P}$, by appending one-vectors to the last dimension. The projective transformation is parameterized by $A \in \mathbf{R}^{4 \times 4}$ and the transformed point $\mathbf{\widetilde{p}}_i \mapsto A \mathbf{\widetilde{p}}_i$. Compared to the affine transformations, projective transformations have more degrees of freedom but cannot preserve parallelism. \textcolor{black}{Projective transformations preserve collinearity and incidence, hence fail to capture all possible deformations. For example, points lying on the same line will always be mapped to a line, and this constraint may be overly restrictive. It is of interest to be able to break this constraint if these points are from different semantic categories. A more general transformation that covers various deformations may be more effective.} %\textcolor{black}{what else does homography preserves?}

\noindent {\bf Deformable Transformations.} When all the points have the freedom to move without much constraint, the 3D shape can deform freely. We refer to this general spatial transformation as a {\it deformable transformation}.  It has more degrees of freedom and does not necessarily preserve the topology. 

Learning new per-point offsets would be computationally hard and costly, we thus use a parametric offset model instead.  Taking both point coordinates and features as inputs, the model would learn offsets dependent upon both spatial and feature representations of the input point cloud.  

\subsection{Spatial Transformers for 3D Point Clouds}
\label{sec:TFLB}

Our so-called \textit{spatial transformer} method applies a geometric transformation to the input point cloud to obtain different local neighborhoods for feature learning.
It can be applied to existing point cloud processing networks as spatial transformers only alter local neighborhoods.  

Suppose at layer $t$, the spatial transformer block contains $k^{(t)}$ transformers. 
Each transformer learns a transformation to apply to the input point coordinates. We refer to the transformed points as nodes of a \textit{sub-graph} and their feature on it the corresponding sub-feature.  We then concatenate all sub-features from these transformers to form the final output of the learning block. Suppose that the $i^\text{th}$ spatial transformer at the $t^{\text{th}}$ layer takes as input the original point cloud $P \in \mathbf{R}^{3 \times N}$ and  previous feature map  $\mathcal{F}^{(t-1)} \in \mathbf{R}^{ \mathfrak{f}^{(t-1)} \times N}$.

\noindent {\bf Affine}. We form $k^{(t)}$ new transformed point from $\mathbf{p}_j$ as:
\begin{equation}
\label{eq:affine}
    \mathbf{g}^{(t)}_{i,j} = A^{(t)}_i \mathbf{p}_j +\mathbf{b}^{(t)}_i, \;\; i = 1, 2, ..., k^{(t)}.
\end{equation}
Since the point affinity matrix is invariant under uniform scaling and translation, we set $\| A_i\|_F =1, b=0$, for all $i$. Thus, with 
 $G^{(t)}_{i} = \left\{ \mathbf{g}^{(t)}_{i,j}\right\}_j$, we
simplify Equation  \ref{eq:affine} as:
\begin{equation}
\label{eq:affine2}
    G^{(t)}_{i} = A^{(t)}_i P , \;\; i = 1, 2, \cdots, k^{(t)}. 
\end{equation}

We compute the $k$ nearest neighbours of each transformed point $ G^{(t)}_{i}$ and obtain the point affinity matrix $S^{(t)}_i$, based on which we define local neighborhoods and apply point convolutions on previous point cloud feature map  $\mathcal{F}^{(t-1)}$.  We get the point cloud feature $ F^{(t)}_i\in \mathbf{R}^{ f^{(t)}_i \times N} $ of the sub-graph from $i$-th transformation and its altered neighborhoods:

\begin{equation}
\label{eq:feature_lear}
    F^{(t)}_i = \textsf{CONV}(\mathcal{F}^{(t-1)} , S^{(t)}_i, k), \;\; i = 1, 2, ..., k^{(t)}, 
\end{equation}
where \textsf{CONV} denotes the point convolution: It takes \begin{inparaenum}[\bfseries (a)] \item previous point cloud features, \item the affinity matrix (for defining local neighborhoods of every point) and \item the number of neighbors (for defining the size of neighborhoods) as inputs. \end{inparaenum} 

In point convolutions such as \cite{DGCNN}, the point affinity matrix changes the input feature in a non-differentiable way. Therefore, we append the transformed point cloud $ P^{(t)}_{i}$ to the input feature for the sake of back-propagating the transformation matrix $A$.  In sampling-based convolutions such as bilateral convolution \cite{su2018splatnet}, the point affinity matrix changes the input feature in a differentiable way; no additional operation is needed.

For all the $k^{(t)}$ sub-graph in layer/ block $t$, we learn $k^{(t)}$ point cloud features $ F^{(t)}_i$.
The output of this module is the concatenation of all the sub-graph point cloud features:
\begin{equation}
     \mathcal{F}^{(t)} = \textsf{CONCAT}(F^{(t)}_1, F^{(t)}_2, ..., F^{(t)}_{k^{(t)}}), 
\label{eq:concat_feat}
\end{equation}
where $F^{(t)}_i\in \mathbf{R}^{ f^{(t)}_i \times N }$ and $\mathfrak{f}^{(t)} = \sum_i^{k^{(t)}} {f}^{(t)}_i$, $\mathcal{F}^{(t)}\in \mathbf{R}^{ \mathfrak{f}^{(t)} \times N}$. 
In our implementation, we randomly initialize $A$ from the standard normal distribution $\mathcal{N}(0,1)$. Before computing the coordinates of the transformed point cloud, we normalize $A$ by its norm $\|A\|_F$, as the point affinity matrix is invariant under uniform scaling. 

\noindent {\bf Projective}. Analogous to the affine spatial transformer, for the $i^{\text{th}}$ graph at $t^{\text{th}}$ layer, we apply a projective transformation to the point cloud $\widetilde{P}$ in homogeneous coordinates and get the transformed point cloud as:
\begin{equation}
\label{eq:proj}
    \widetilde{G}^{(t)}_{i} = B^{(t)}_i \widetilde{P} , \;\; i = 1, 2, \cdots, k^{(t)}, 
\end{equation}
where $B^{(t)}_i \in \mathbf{R}^{4 \times 4}$ is the transformation matrix in  homogeneous coordinates.
We then follow the same procedure as in Equations \ref{eq:feature_lear} and \ref{eq:concat_feat} to get the output feature $\mathcal{F}^{t}$.

\noindent {\bf Deformable.} Affine and projective transformations can transform the input point cloud, alter the point affinity matrix, and provide learnable local neighborhoods for point convolutions at different layers. 
However, they are limited as affine transformations are linear and projective transformations map lines to lines only.  We define a non-linear deformable spatial transformer at the $t^{\text{th}}$ layer and $i^{\text{th}}$ sub-graph as
\begin{equation}
\label{eq:deform}
G^{(t)}_{i} = A^{(t)}_{i} P + D^{(t)}_{i},
\end{equation}
where $A^{(t)}_{i} P$ is the  affine transformation component and  $D^{(t)}_{i}\in \mathbf{R}^{3 \times N}$ gives every point additional freedom to move, so  the point cloud has the flexibility to deform its shape.  Note that the translation vector $\mathbf{b}$ in Equation \ref{eq:affine} is a special case of the deformation matrix $D^{(t)}_{i}$.  In general, the deformation matrix $D^{(t)}_{i}$ can significantly change local neighborhoods.

The spatial transformer parameters are learned in an end-to-end fashion from both point cloud coordinates and features.  Since affine transformation $A^{(t)}_{i} P$ is dependent on spatial locations, we let the deformation matrix $D^{(t)}_{i}$  depend on the features: $D^{(t)}_{i} = \mathcal{C}^{(t)}_{i} \mathcal{F}^{(t-1)}$, where $\mathcal{C}^{(t)}_{i} \in \mathbf{R}^{3\times \mathfrak{f} } $ transforms the previous layer feature  $\mathcal{F}^{(t-1)} \in \mathbf{R}^{\mathfrak{f} \times N}$ from $\mathbf{R}^{\mathfrak{f}}$ to $\mathbf{R}^3$.  Hence, the deformable transformation in Equation \ref{eq:deform} can thus be simplified as:
\begin{equation}
\label{eq:deform2}
    G^{(t)}_{i} = \begin{bmatrix} 
A^{(t)}_{i} & \mathcal{C}^{(t)}_{i}
\end{bmatrix} 
\begin{bmatrix} 
P \\
\mathcal{F}^{(t-1)}
\end{bmatrix} = C^{(t)}_{i} \begin{bmatrix} 
P \\
\mathcal{F}^{(t-1)}
\end{bmatrix},
\end{equation}
where $C^{(t)}_i \in \mathbf{R}^{3 \times (3+f^{(t-1)} )}$ is the concatenation of affine and deformable transformation matrix that captures both point cloud coordinates and features.

After we compute the transformed point coordinates $G^{(t)}$, we follow Equations \ref{eq:feature_lear} and \ref{eq:concat_feat} to learn the feature of each transformed sub-graph and concatenate them as the final output feature of layer $t$.

Our deformable spatial transformer has two parts: $A^{(t)}_{i} P$ and $\mathcal{C}^{(t)}_{i} \mathcal{F}^{(t-1)}$, for a linear transformation of 3D spatial coordinates and a nonlinear transformation of point features (which reflect semantics) respectively.  In Section \ref{sec:ablation}, we provide empirical analysis of these two components.

\subsection{Spatial Transformer Networks}

We spatially transform the input point cloud in order to obtain dynamic local neighborhoods for point convolutions. 
The transformer can be easily added to existing point cloud processing networks. We first describe the procedure and then provide three applications with several networks.

\begin{figure*}[!ht]
  \centering
  \includegraphics[width=0.85\textwidth]{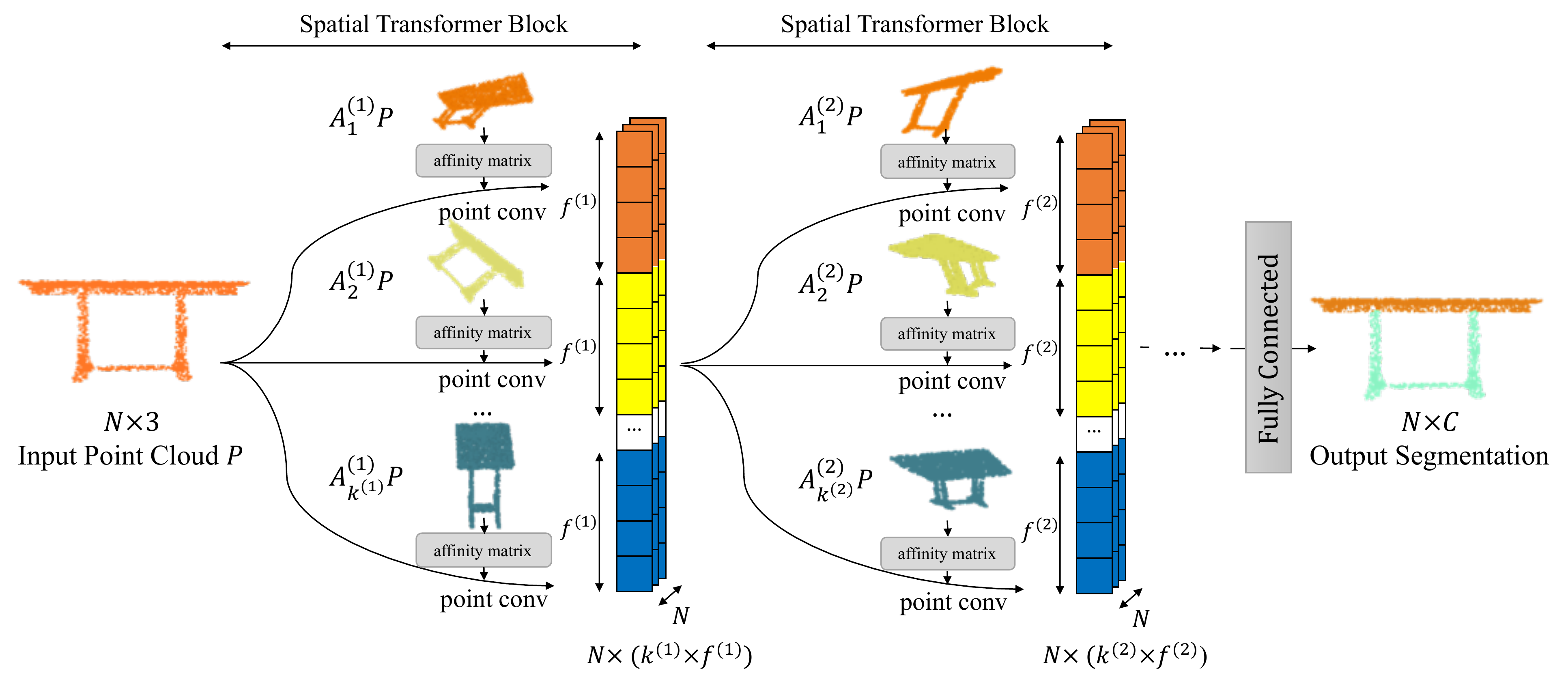}
  \caption{  The point cloud segmentation network with spatial transformers. Our network consists of several spatial transformers. At each layer, we learn  $k$ transformation matrices $A$ to apply to the input point cloud  $P$, and compute the corresponding point affinity matrices based on their $k$-NN graphs. For each sub transformation, we can learn a sub-feature, and then concatenate all features to form an output feature of dimension $f \times N$. The output feature will be used for the next spatial transformer block for feature learning. By stacking several such transformation learning blocks and finally a fully connected layer of dimension $C$ (the number of class), we can map the input point cloud to the $C \times N$ segmentation map.}
  \label{fig:network}
  \vspace{-1em}
\end{figure*}

\noindent {\bf Point Cloud Networks with Spatial Transformers.} 
Consider segmenting  $N$ 3D points into  $C$ classes as an example.
Fig.\ref{fig:network} depicts a general network architecture for point cloud segmentation, where several spatial transformers are used at different layers.
At layer $t$, we learn  $k^{(t)}$ transformation matrices $\{A^{(t)}_i\}_{i=1}^{k^{(t)}}$, apply each to the input point coordinates $P$, and then compute the point affinity matrices $\{S^{(t)}_i\}_{i=1}^{k^{(t)}}$, e.g., based on $k$-NN graphs for the edge convolution \cite{DGCNN}. 

For each sub-transformation, we learn a feature $F^{(t)}_i$ of dimension $N \times {f}^{(t)}_i$.  We then concatenate all $k^{(t)}$ features at this layer to form an output feature $\mathcal{F}^{t}$ of dimension $N \times \mathfrak{f}^{(t)}$, where $\mathfrak{f}^{(t)} = \sum_i^{k^{(t)}} {f}^{(t)}_i$.  The output feature serves as the input to the next layer for further feature learning. 

Note that affine or projective transformation matrices are applied to the original point cloud coordinates $P$, since each layer has not just one but  multiple spatial transformers.  However, the deformable transformation matrix $\mathcal{C}_i^{(t)}$ is applied to the previous feature map, the feature transformation component is thus progressively learned. 

\begin{figure}[!t]
  \centering
  \includegraphics[width=0.5\textwidth]{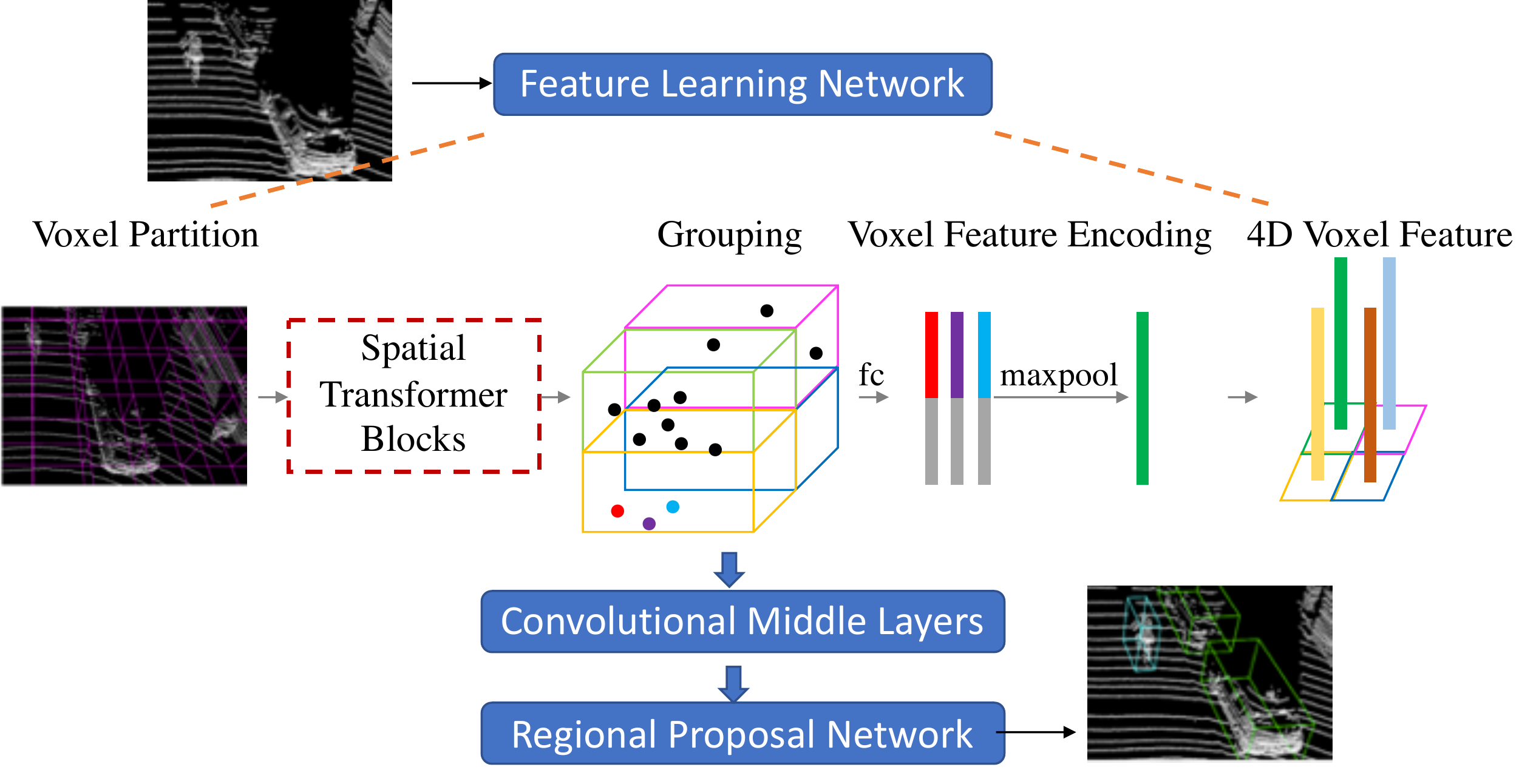}
  \caption{ The object detection network. 
  We add spatial transformers to the the point feature learning network of
 \cite{voxelnet} for obtaining dynamic local neighborhoods.
 Transformers only affect feature learning but not point coordinates for grouping.
 }
  \label{fig:detection_net}

\end{figure}

By stacking several such transformation learning blocks and finally a fully connected layer of dimension $C$, we can map the input point cloud to the segmentation map of dimension $C \times N$, or downsample to a vector of dimension $C$ for classification tasks. 
For the spatial transformer block in a point cloud detection network (Fig.\ref{fig:detection_net}),  $C$ is the dimension of the output feature.
We train the network end-to-end. 

\noindent \textbf{Classification Networks.} A point cloud classifier \cite{pointnet++, li2018pointcnn} takes 3D points, learns features from their local neighborhoods, and outputs $C$ classification scores, where $C$ is the number of classes.  We add spatial transformers at each layer to obtain different local neighborhoods for feature learning. 

\noindent \textbf{Point-based Segmentation Networks.}
These networks \cite{pointnet++, pointnet, li2018pointcnn, DGCNN} take 3D points and compute their point affinity matrices and local neighborhoods from the point coordinates. Features are learned by applying  convolution operators on the points and their local neighborhoods.

We use the \textit{edge convolution} in \cite{DGCNN} as our baseline, which takes relative point coordinates as inputs and achieves the state-of-the-art performance. Specifically, we retain their learning settings and simply insert  spatial transformers to generate new local neighborhoods for the edge convolutions.

\noindent \textbf{Sampling-based Segmentation Networks.}
To demonstrate the general applicability of our spatial transformers, we consider point affinity matrices on transformed point clouds as defined in sampling-based networks such as SplatNet \cite{su2018splatnet}.  

SplatNet groups 3D points onto a permutohedral lattice \cite{adams2010fast} and applies bilateral filters \cite{bilaterCNN} on the grouped points to get features. The permutohedral lattice defines the local neighborhoods of every point and makes the bilateral convolution possible.  We add spatial transformers to deform the point cloud and form various new lattices.  The local neighborhoods can dynamically configure for learning point cloud semantics.  We keep all the other settings of SplatNet.

\noindent \textbf{Detection Networks.} Detecting objects in a 3D point cloud generated from e.g. LiDAR sensors is important for autonomous navigation, housekeeping robots, and AR/ VR.  These 3D points are often sparse and imbalanced across semantic classes.  Our spatial transformers can be added to a detection network and improve feature learning efficiency and task performance with dynamic local neighborhoods.

Our baseline is VoxelNet \cite{voxelnet},  the state-of-the-art 3D object detector for autonomous driving data.  We adopt all its settings, and add spatial transformers on the raw point cloud data, before point grouping (Fig.\ref{fig:detection_net}). 
To demonstrate that spatial transformers enhance feature learning for point cloud processing, we let transformers only affect point features but not point coordinates for grouping.  With spatial transformers, point coordinates could also be transformed at the grouping stage, which would lead to non-cuboid 3D detection boxes.  Although interesting, we do not explore this variation and deem it beyond the scope of this paper. 

\subsection{Relevance to Other Works}
\label{sec:relevance}

We review related works on deformable convolutions \cite{dai2017deformable,thomas2019kpconv} and DGCNN \cite{DGCNN}.

\noindent \textbf{Deformable Convolutions.}
Deformable convolutional networks \cite{dai2017deformable} learn dynamic local neighborhoods for 2D images. Specifically, at each  location $\mathbf{p_0}$ of the output feature map $Y$, deformable convolutions modify the regular grid $R$ with offsets  $\{\Delta \mathbf{p_n}\}_{n=1}^N$, where $N = |R|$. The output on input $X$ by convolution with weight $\mathbf{w}$ becomes:
\begin{equation}
Y(\mathbf{p_0}) = \sum_{\mathbf{p_n} \in \mathbf{R}^3} w(\mathbf{p_n}) X (\mathbf{p_0} + \mathbf{p_n} + \Delta \mathbf{p_n})
\end{equation}
Note that KPConv \cite{thomas2019kpconv} directly adapts this formula to point clouds as deformable point convolutions. 
Although also achieving dynamic local neighborhoods, our spatial transformers alter neighborhoods differently:
\begin{enumerate}[leftmargin=*] 
\item Deformable point convolutions learn to alter each neighborhood with an offset to the regular grid $R$.  We learn global transformations on the input point cloud and the metric of defining local neighborhoods changes.  Global transformations like affine transformations can retain the global geometric properties such as collinearity and parallelism, while local transformations has no such constraints as only local neighborhoods are available. 
\item Offsets of deformable point convolutions are dependent upon feature values, while transformation matrices of our spatial transformers are dependent upon point coordinates for affine and projective transformations, or both point coordinates and feature values for deformable transformations.  Access to point coordinates provides additional information and regularzation. 
\end{enumerate}

\noindent \textbf{Dynamic Graph CNN.}
Dynamic local neighborhoods have also been explored in DGCNN \cite{DGCNN} for point cloud processing.  It has three main differences with our work. 
\begin{enumerate}[leftmargin=*] 
\item How neighborhoods are defined is different. DGCNN uses high-dimensional feature maps to construct the point affinity matrix and generate local neighborhoods. 
Our local neighborhoods are from transformed point clouds.
Reusing point features for defining neighborhoods may be straightforward, but reduce the distinction between spatial and semantic information and hurt generalization.
\item It is computationally costly to build dense nearest neighbor graphs in a high-dimensional feature space. 
\item DGCNN \cite{DGCNN} uses only one nearest neighbor graph at different layers, whereas we have multiple graphs at each layer for capturing different geometric transformations.
\end{enumerate}
 With less computational cost and more flexibility in geometric transformations, we achieve better empirical performance on semantic segmentation (Table \ref{tab:partseg} and Table \ref{tab:semseg}).

%% file: sections/4results.tex
\section{Experiments}

We conduct comprehensive experiments to verify the effectiveness of our spatial transformers. 
We benchmark with two types of networks, point-based and sampling-based metrics for defining point neighborhoods, on four point cloud processing tasks: classification, part segmentation, semantic segmentation and detection. 
We conduct ablation studies on deformable spatial transformers. We further provide visualization, analysis and insights of our method.
\subsection{Classification}
\label{sec:classM}

\begin{table*}[!ht]
\begin{center}

\centering
\caption{\small Spatial transformers improves ModelNet40 classification accuracy. We report ModelNet40 classification accuracy of different baselines, without and with spatial transformers, with or without random rotations.  \textit{Point-based} refers to the baseline method adopting Euclidean-distance based affinity matrices \cite{DGCNN}. \textit{Sampling-based} refers to the baseline method adopting permutohedral-lattice based affinity matrices \cite{su2018splatnet}. We observe accuracy gains for different baseline networks with spatial transformers. Transformer gains are invariant to input rotations.}

\vspace{-1.3em}

\label{tab:cls}
\resizebox{1\textwidth}{!}{
\begin{tabular}{>{\columncolor[gray]{0.95}}c|cc|cccc|cc|cccc}
\hline
\rowcolor{LightCyan}
\multirow{2}{*}{} & & & \multicolumn{4}{c|}{Point-based} & \multicolumn{2}{c|}{Point-based with rand. input rotations} & \multicolumn{4}{c}{Samplin-based } \\ \cline{4-11} 
 \rowcolor{LightCyan2}                 &          PointNet \cite{pointnet}                &         DGCNN \cite{DGCNN}            &  \cite{DGCNN}  (fixed)  & Affine & \textcolor{black}{Proj.}   & Deformable   &  \cite{DGCNN}  (fixed)           & Deformable              &  SplatNet\cite{su2018splatnet}  & Affine    &\textcolor{black}{Proj.} & Deformable    \\ \hline
Avg.         & 86.2                 & 89.2                & 88.8    & 89.3 & \textcolor{black}{89.2} & \textbf{89.9}  & 85.7               & \textbf{88.3}             & 86.3     & 87.4 & \textcolor{black}{87.1} & \textbf{88.6}  \\ \hline

\end{tabular}
}
\end{center}
\end{table*}

\begin{figure}[!ht]
\centering
        \includegraphics[width=0.45\textwidth]{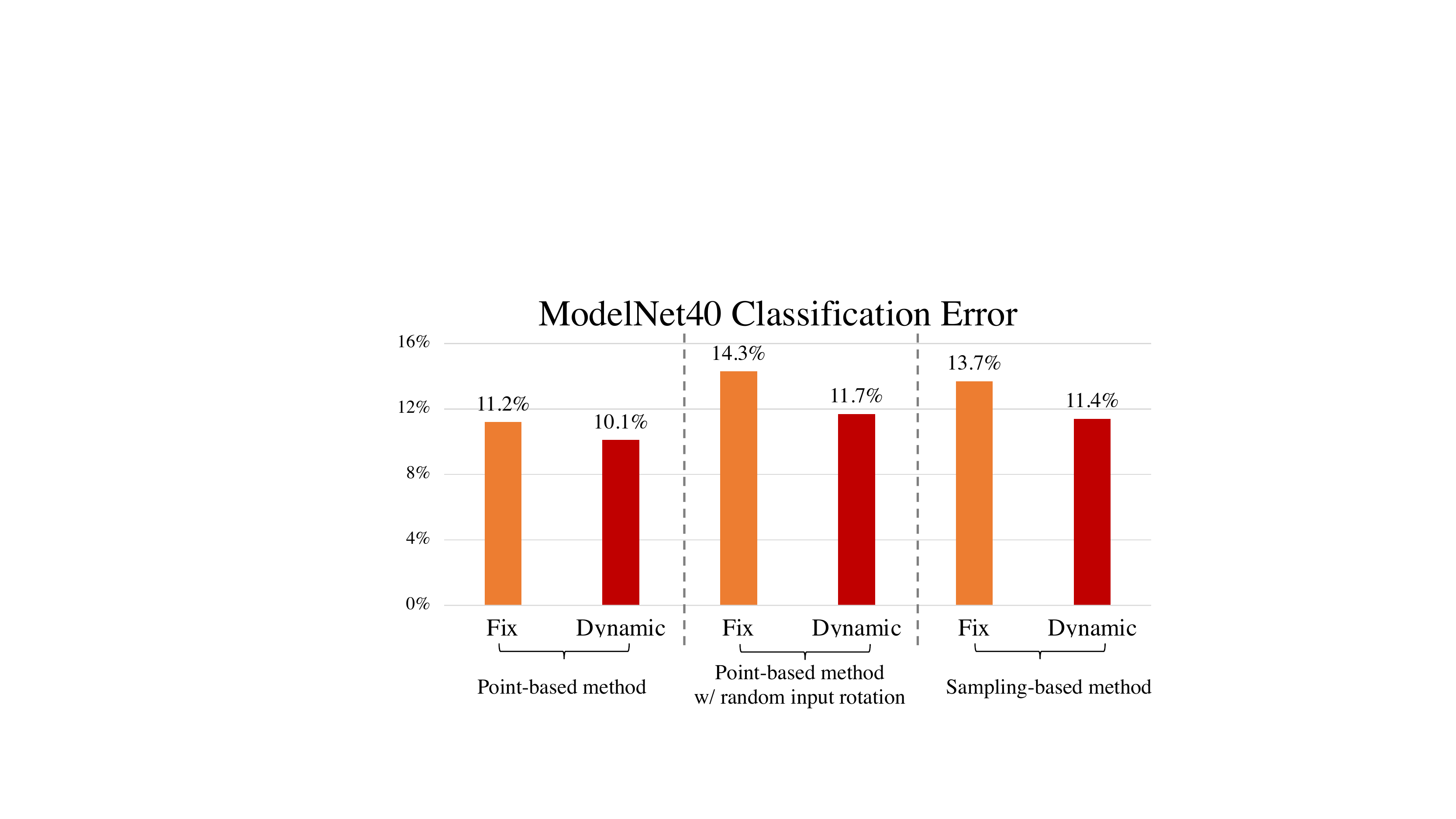}%
        \begin{flushleft}
        \caption{Spatial transformers lead to higher accuracy and more rotation invariance on ModelNet40. We report classification errors for different baselines, without and with spatial transformers, with or without random rotations. Transformers consistently lead to the lower errors than fixed graph baselines, and the improvement is larger upon random rotations.}
        \end{flushleft}
   \label{fig:class}
   \end{figure}

We benchmark on ModelNet40 3D shape classification \cite{wu20153d}. We add transformers to two baselines \cite{DGCNN, su2018splatnet} and adopt the same network architecture, experimental setting and evaluation protocols.  
Table \ref{tab:cls} and Fig.\ref{fig:class} show that adding spatial transformers to point-based and sampling-based method gives 1\% and 2\% gain.  

In addition, our performance gain over \cite{DGCNN}, which builds one per-layer dynamic neighborhood graphs with high-dimensional point features, demonstrates the advantages of our method of building multiple dynamic neighborhood graphs with transformed 3D point coordinates.

Fig.\ref{fig:part_deform} shows that
spatial transformers align the 3D shape better according to its semantics.  We augment training and testing data with random rotations, and observe that spatial transformers gain 3\% over its fixed graph counterpart. 

\subsection{Part Segmentation}
\label{sec:partseg_p}
We benchmark on ShapeNet part segmentation \cite{chang2015shapenet},
where the goal is to assign a part category label (e.g. chair leg, cup handle) to each 3D point.
The dataset contains $16,881$ shapes from $16$ categories, annotated with $50$ parts in total, and the number of parts per category ranges from $2$ to $6$.

\begin{figure}[!t]
\centering
        \includegraphics[width=0.5\textwidth]{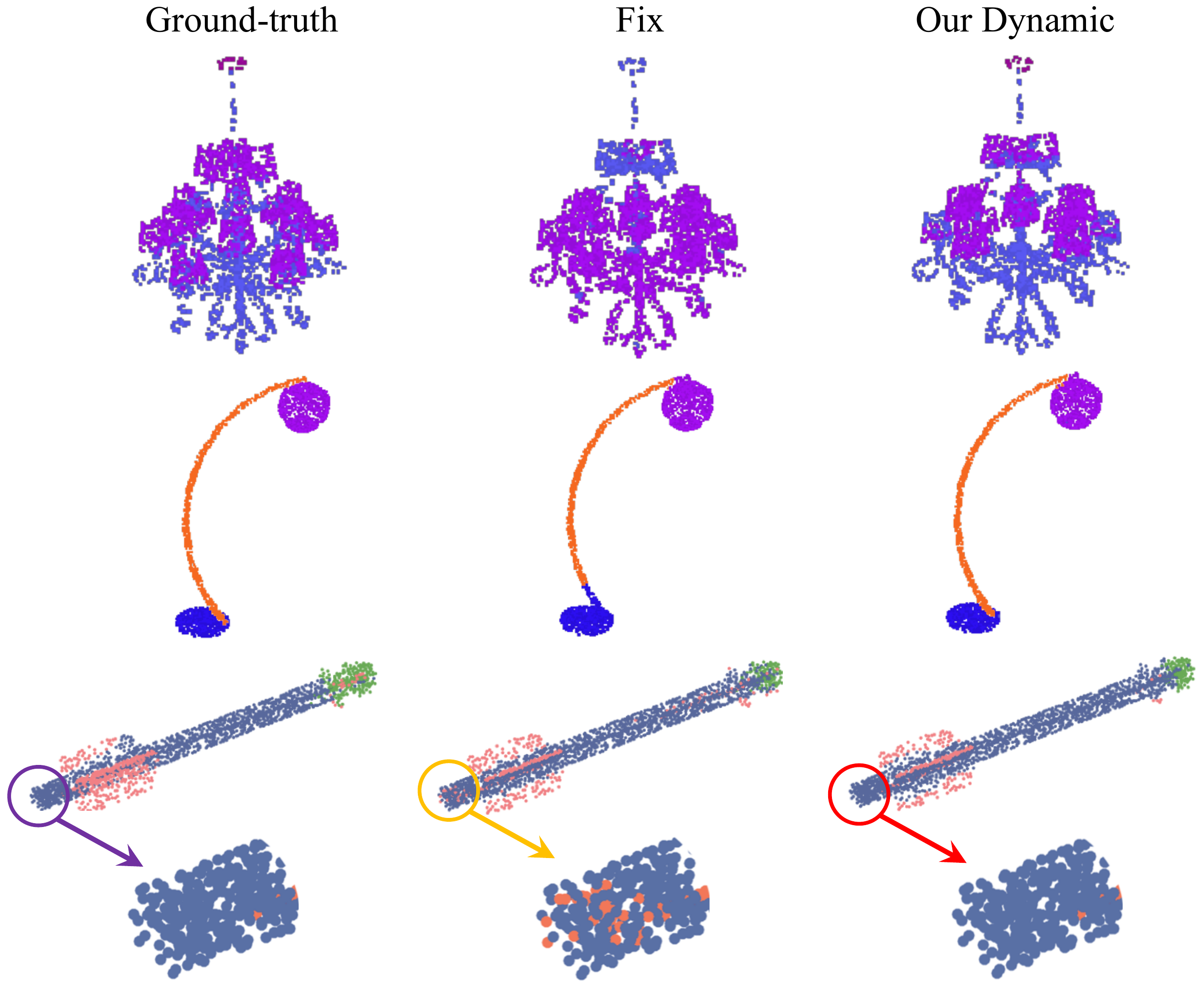}%
\caption{ Spatial transformers improve the part segmentation performance.
We show part segmentation results of different baselines, where different parts are marked with different colors.
With spatial transformers, part segmentation for objects with less rigid and more complicated structures improves (1st and 2nd row, lamp). The segmentation consistency within each part also improves (3rd row, rocket).
}
   \label{fig:partseg}
   \end{figure}

\subsubsection{Point-based Method} 

\textbf{Network Architectures.} 
\textit{Point-based} methods  construct neighborhoods based on point coordinate operations such as \textit{edge convolution} for our baseline DGCNN \cite{DGCNN}.
We follow the same network architecture and evaluation protocols of  \cite{DGCNN}.  The network has 3 convolutional layers; the output feature dimension is 64.  To capture information at different levels,  all the convolutional features are concatenated and fed through several fully connected layers to output the segmentation.

As a \textit{fixed  graph} baseline, we use the same input point coordinates as the metric to define fixed local neighborhoods. 
We insert spatial transformers to alter the metric for defining point neighborhoods for edge convolutions. 
There are point-based \textit{affine}, \textit{projective} and \textit{deformable} networks when inserting different spatial transformers (Section \ref{sec:TFLB}). 
As for classification, \cite{DGCNN} directly uses learned features to build  point affinity matrices for dynamic neighborhoods.

We follow \cite{DGCNN} and use three edge convolution layers.
At each layer,  we keep the number of graphs  $k$ and sub-graph feature dimension $f$ the same, and search for the best architecture.
We report results of affine, projective and deformable networks with $k\!=\!4$, $f\!=!32$.
For fair comparisons, we increase the number of channels of baselines so all the methods have the same number of parameters.

\noindent \textbf{Results and Analyses}. In Table \ref{tab:partseg}, we report the instance average mIOU (mean intersection over union), as well as the mIOU of some representative categories in ShapeNet.  Compared with the fixed graph baseline, the affine, projective and deformable spatial transformers achieve 0.5\%, 0.2\% and 1.1\% improvement respectively  and beats the fixed graph baseline methods in most categories.  Specifically, we observe 8.0\%, 8.3\% and 4.7\% performance boost with spatial transformers over the fixed graph baseline.  Our deformable spatial transformers gain 4.0\% over \cite{DGCNN}.

We also beat other state-of-the-art methods \cite{pointnet,pointnet++,rethage2018fully} by a significant margin. Adding deformable spatial transformers to PointCNN \cite{li2018pointcnn} gains 6\% (4\%) on motorbike (bag) and 1\% on average. 
We observe that categories with fewer samples are more likely to gain  possibly due to regularization by transformers.  
Fig.\ref{fig:partseg} shows that deformable spatial transformers make   more smooth predictions and achieve better performance than the fixed graph baseline.

\begin{figure}[!t]
  \centering
  \includegraphics[width=0.5\textwidth]{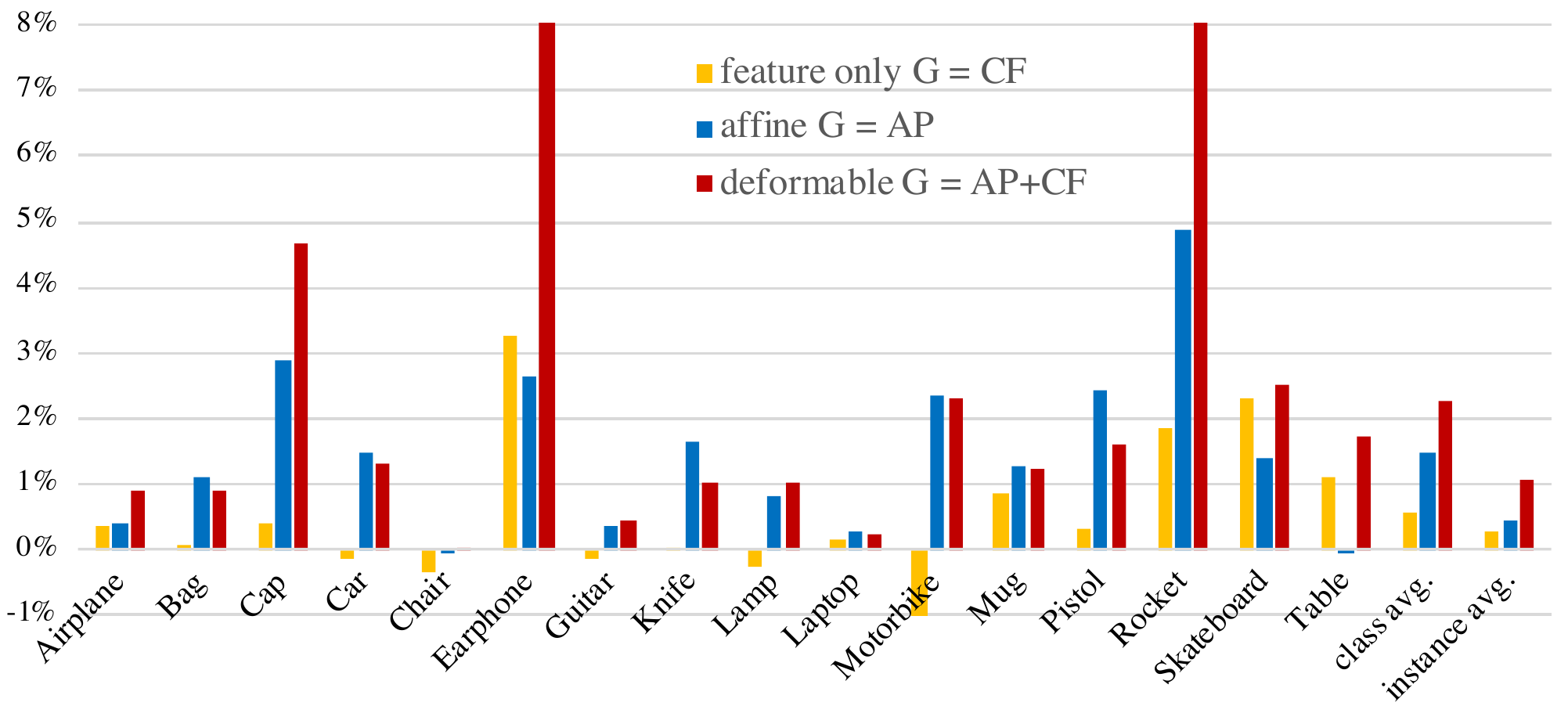}
  \caption{
  Transforming both point cloud coordinates and features for dynamic local neighborhoods leads to the largest gain.
  We report different parts of deformable transformers' performance gain over fixed local neighborhood baseline on ShapeNet part segmentation. 0\% means achieving the same accuracy as fixed local neighborhood baseline and negative value means achieving worse accuracy than fixed neighborhood baseline. 
Compared with  preserving either affine part $AP$ or feature part $CF$,  deformable spatial transformers ($AP+CF$) achieves largest gains on every category, specifically 8\% gains on earphone and rocket.
  }
  \label{fig:deform_ablation}
    \vspace{-1em}
\end{figure}

From affine to deformable transformations, the performance increases as the degree of freedom increases for the transformer.
Projective transformers, however, perform slightly worse than affine transformers.  The performance drop could result from geometrical distortion caused by mapping 3D points with homogeneous coordinates.
Furthermore, for deformable transformers, when removing the constraint that the transformed points should be similar to the input point cloud (Fig.\ref{fig:deform_ablation}, feature only $G=CF$), the performance also drops, indicating the necessity of the proposed similar-to-input constraint on spatial transformers.

\begin{table*}[]
\begin{center}
\LARGE

\caption{Spatial transformers improve part segmentation performance. We report mIoU(\%) on ShapeNet PartSeg dataset. Compared with several other methods, deformable spatial transformers achieve the SOTA in average mIoU.}

\vspace{-1.3em}
\label{tab:partseg}
\resizebox{1\textwidth}{!}{

\begin{tabular}{>{\columncolor[gray]{0.95}}l|c|ccccccccccccccccc} 
\hline
\rowcolor{LightCyan}

                                       & Avg.          & aero          & bag           & cap           & car      & chair          & earphone & guitar & knife & lamp & laptop & motorbike & mug & pistol & rocket & skateboard & table         \\ 
\hline
\rowcolor{LightCyan2} \# shapes &              & 2690& 76& 55& 898& 3758& 69& 787& 392& 1547& 451& 202& 184& 283& 66& 152& 5271\\ 
\hline
3DCNN \cite{pointnet}                         & 79.4& 75.1& 72.8& 73.3& 70.0& 87.2& 63.5& 88.4& 79.6& 74.4& 93.9& 58.7& 91.8& 76.4& 51.2& 65.3& 77.1 \\
PointNet\cite{pointnet}                                & 83.7& 83.4& 78.7& 82.5& 74.9& 89.6& 73.0& 91.5& 85.9& 80.8& 95.3& 65.2& 93.0& 81.2& 57.9& 72.8& 80.6   \\
PointNet++ \cite{pointnet++}                            & 85.0& 82.4& 79.0& 87.7& 77.3& 90.8& 71.8& 91.0& 85.9& 83.7& 95.3& 71.6& 94.1& 81.3& 58.7& 76.4& \textbf{82.6}      \\
FCPN   \cite{rethage2018fully}                                & 81.3& 84.0& 82.8& 86.4& \textbf{88.3}& 83.3& 73.6& \textbf{93.4}& 87.4& 77.4& \textbf{97.7}& \textbf{81.4}& 95.8& \textbf{87.7}& \textbf{68.4}& \textbf{83.6}& 73.4   \\
DGCNN  \cite{DGCNN}             & 81.3& 84.0& 82.8& 86.4& 78.0& \textbf{90.9}& 76.8& 91.1& 87.4& 83.0& 95.7& 66.2& 94.7& 80.3& 58.7& 74.2& 80.1 \\ 
\hline
Point-based \cite{DGCNN} fixed graph                   & 84.2 & 83.7 & 82.4 & 84.0   & 78.2 & \textbf{90.9} & 69.9 & 91.3 & 86.6 & 82.5 & 95.8 & 66.5 & 94.0   & 80.8 & 56.0   & 73.8 & 79.8 \\
Point-based affine                     & 84.7 & 84.1 & 83.5 & 86.9 & 79.6 & \textbf{90.9} & 72.5 & 91.6 & 88.2 & 83.3 & 96.1 & 68.9 & 95.3 & 83.3 & 60.9 & 75.2 & 79.7 \\
Point-based projective                 & 84.4 & 84.3 & 84.2 & 88.5 & 77.9 & 90.4 & 72.8 & 91.2 & 86.6 & 81.7 & 96.0   & 66.6 & 94.8 & 81.3 & 61.6 & 72.1 & 80.5  \\
Point-based deformable                 & 85.3 & \textbf{84.6} & 83.3 & 88.7 & 79.4 & \textbf{90.9} & 77.9 & 91.7 & 87.6 & 83.5 & 96.0   & 68.8 & 95.2 & 82.4 & 64.3 & 76.3 & 81.5     \\ 

\textcolor{black}{Point-based deformable random}              & \textcolor{black}{84.7} & \textcolor{black}{84.3} & \textcolor{black}{84.4} & \textcolor{black}{83.2} & \textcolor{black}{78.9} & \textcolor{black}{90.8} & \textcolor{black}{75.6}   & \textcolor{black}{91.4} & \textcolor{black}{87.1} & \textcolor{black}{83.0} & \textcolor{black}{95.9} & \textcolor{black}{66.8} & \textcolor{black}{94.8} & \textcolor{black}{82.1} & \textcolor{black}{62.3} & \textcolor{black}{75.7} & \textcolor{black}{80.4}       \\\hline
PointCNN \cite{li2018pointcnn} & 84.9 & 82.7 & 82.8 & 82.5 & 80.0   & 90.1 & 75.8 & 91.3 & \textbf{87.8} & 82.6 & 95.7 & 69.8 & 93.6 & 81.1 & 61.5 & 80.1 & 81.9\\
PointCNN deformable & \textbf{85.8} & 83.4 & \textbf{86.6} & 85.5 & 79.1 & 90.3 & \textbf{78.5} & 91.6 &\textbf{87.8} & 84.2 & 95.8 & 75.3 & 94.6 & 83.3 & 65.0   & 80.7 & 81.7 \\ 
\hline
Sampling-based baseline \cite{su2018splatnet}               & 84.6 & 81.9 & 83.9 & 88.6 & 79.5 & 90.1 & 73.5 & 91.3 & 84.7 & 84.5 & 96.3 & 69.7 & 95.0   & 81.7 & 59.2 & 70.4 & 81.3      \\
\textcolor{black}{Sampling-based projective}              & \textcolor{black}{84.4} & \textcolor{black}{82.1} & \textcolor{black}{84.0} & \textcolor{black}{\textbf{89.1}} & \textcolor{black}{77.9} & \textcolor{black}{89.6} & \textcolor{black}{73.7}   & \textcolor{black}{91.1} & \textcolor{black}{83.3} & \textcolor{black}{83.0} & \textcolor{black}{96.3} & \textcolor{black}{67.2} & \textcolor{black}{94.5} & \textcolor{black}{79.8} & \textcolor{black}{60.0} & \textcolor{black}{68.8} & \textcolor{black}{82.1}       \\
Sampling-based deformable              & 85.2 & 82.9 & 83.8 & 87.6 & 79.6 & 90.6 & 73.0   & 92.2 & 86.1 & \textbf{85.7} & 96.3 & 72.7 & \textbf{95.8} & 83.1 & 65.1 & 76.5 & 81.3       \\
\hline
\end{tabular}
} 
\vspace{-1em}
\end{center}
\end{table*}

\label{sec:defrom_exp}

\subsubsection{Sampling-based Method} 
\label{sec:partseg_s}

\noindent \textbf{Network Architectures.} 
\textit{Sampling-based} methods construct neighborhoods are based on sampling operations on point coordinates.
SplatNet \cite{su2018splatnet} groups points on permutohedral lattices and applies learned bilateral filters \cite{bilaterCNN} on naturally defined local neighbors to extract features. 
We follow the same architecture as SplatNet \cite{su2018splatnet}.  The network starts with a single $1\times 1$ regular convolutional layer, followed by 5 bilateral convolution layers (BCL). The output of all BCL are concatenated and fed to a final $1\times 1$ regular convolutional layer to get the segmentation output. Since each BCL directly takes raw point locations, we consider it as a fixed graph baseline. We add deformable spatial transformers to the network and feed transformed point graphs to BCL to construct permutohedral lattices. With gradients on the permutohedral lattice grid, we can make the transformation matrix learned end-to-end.  Note that we increase the channel of convolution layers for fair comparisons.

\noindent \textbf{Results and Analyses.}
Table \ref{tab:partseg} shows that our deformable spatial transformers (with $k=1$ at all BCLs) gains over the sampling-based fixed graph baseline \cite{su2018splatnet} in most categories with 0.6\% on average and  5.9\% for the rocket category.  It also beats other state-of-the-art baselines.

\subsection{Semantic Segmentation}

\label{sec:semseg}
\begin{figure*}[!ht]
  \centering
      \includegraphics[scale=0.75]{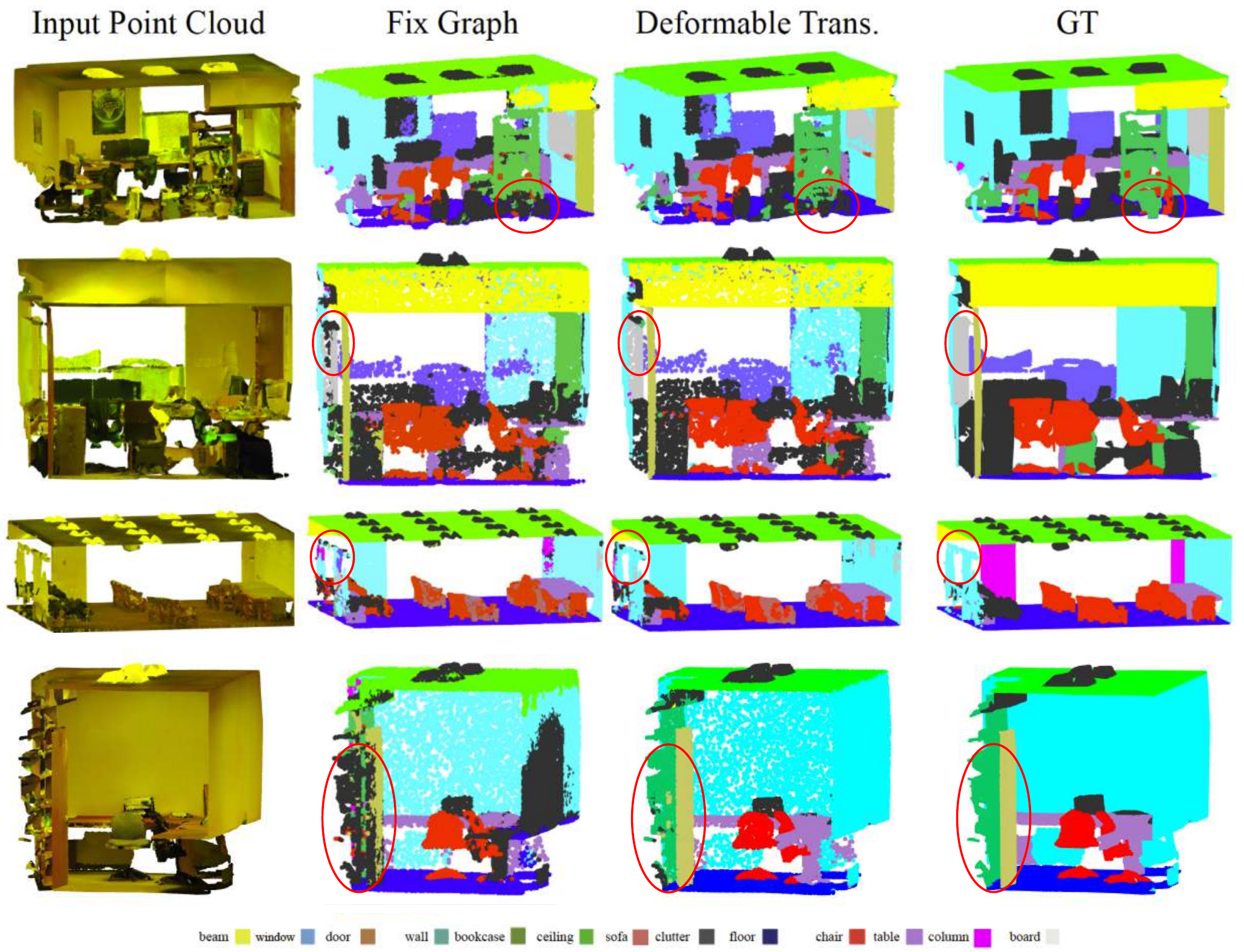}
 \caption{ Spatial transformers improve semantic segmentation results.
 We show qualitative visualizations for semantic segmentation of  deformable spatial transformers and the fixed local neighborhood baseline. The first column is the input point cloud, the second and the third column shows the fixed graph and our spatial transformer results, and the last column is the ground truth.
  Points belonging to different semantic regions are colored differently. We observe better and more consistent segmentation result with our spatial transformer, specifically for the areas circled in red.
 }
  
  \label{fig:sem_result}
\end{figure*}

We benchmark on the Stanford 3D semantic parsing dataset \cite{armeni20163d}.  It contains 3D scans by Matterport covering 6 areas and 271 rooms. Each point is annotated into one of 13 categories such as {\it chair, table, floor,  clutter}.  We follow the data processing procedure of \cite{pointnet}: We first split points by room, and then sample rooms into several 1m $\times $1m blocks. When training, 4096 points are sampled from the block on the fly. We train our network to predict the point class in each block, where each point is represented by 9 values: XYZ, RGB and its [0,1]-normalized location with respect to the room. 

\begin{table}[]
\begin{center}
\LARGE

\caption{
Spatial transformers improve  semantic segmentation performance. We report mIoU(\%) on S3DIS semantic segmentation dataset.
Adding spatial transformers to \cite{DGCNN} and \cite{su2018splatnet} improves the performance.
}
\vspace{-1.3em}
\label{tab:semseg}
\centering
\resizebox{0.5\textwidth}{!}{

\begin{tabular}{>{\columncolor[gray]{0.95}}c|ccccccc}
\hline
\rowcolor{LightCyan}
            & PointNet\cite{pointnet}      & DGCNN\cite{DGCNN}           & \cite{DGCNN}(FIXED)     & \cite{DGCNN}+AFF        & \cite{DGCNN}+DEF   &  SplatNet \cite{su2018splatnet}              &    \cite{su2018splatnet}+DEF            \\\hline
          & 47.7          & 56.1           & 56.0          & 56.9          & \textbf{57.2} &  54.1     &      55.5    \\ \hline
   \rowcolor{LightCyan}
            & ceiling       & floor & wall          & beam          & column        & window & clutter        \\ \hline
\cite{DGCNN}(FIXED)   & 92.5          & 93.1           & 76.1          & 51.0          & \textbf{41.7} & \textbf{49.6}  & 46.8           \\
\cite{DGCNN}+AFF      & 92.7          & \textbf{93.6}  & 76.7          & 52.6          & 41.2          & 48.7           & 47.8           \\
\textcolor{black}{\cite{DGCNN}+PROJ}      & \textcolor{black}{92.5}          & \textcolor{black}{93.5}  & \textcolor{black}{76.7}          & \textcolor{black}{52.7}          & \textcolor{black}{40.7}          & \textcolor{black}{48.5}           & \textcolor{black}{\textbf{48.0}}           \\
\cite{DGCNN}+DEF & \textbf{92.8} & \textbf{93.6}  & \textbf{76.8} & \textbf{52.9} & 41.1          & 49.0           & \textbf{48.0}  \\ \hline
\rowcolor{LightCyan}
            & door          & table          & chair         & sofa          & bookcase      & board          &                \\ \hline
\cite{DGCNN}(FIXED)   & 63.4          & 61.8           & 43.1          & 23.3          & \textbf{42.0} & 43.5           &                \\
\cite{DGCNN}+AFF     & \textbf{63.7} & 63.4           & 45.1          & 27.0          & 41.3          & 44.8           &                \\
\textcolor{black}{\cite{DGCNN}+PROJ}     & \textcolor{black}{\textbf{63.5}} & \textcolor{black}{62.3}          & \textcolor{black}{44.8}          & \textcolor{black}{27.0}          & \textcolor{black}{41.5}         & \textcolor{black}{44.9}           &                \\
\cite{DGCNN}+DEF  & 63.5          & \textbf{64.2}  & \textbf{45.2} & \textbf{28.1} & 41.7          & \textbf{46.1}  & \\\hline          

\end{tabular}
}\vspace{-0.7em}

\end{center}
\end{table}

\noindent \textbf{Network Architectures}. We adopt DGCNN \cite{DGCNN}
as Section \ref{sec:partseg_p}, with $C=13$, the number of semantic categories.

\noindent \textbf{Results and Analyses}.  In terms of average mIoU, Table \ref{tab:semseg} shows that affine and deformable spatial transformers gain  0.9\% and 1.2\% respectively over the fixed graph baseline.  Deformable transformers also gain 1.1\% over \cite{DGCNN} and beat all other state-of-the-art methods.  Likewise for sampling-based methods \cite{su2018splatnet}, we observe 1.4\% gain.

As for part segmentation, semantic segmentation performance improves when point clouds are given more freedom to deform (from affine to deformable spatial transformers) based on transformation of original locations and feature projections.  Projective transformers give least performance gain,  suggesting that mapping 3D points via homogeneous coordinates may not be most efficient. 

 Fig.\ref{fig:sem_result} shows that semantic segmentation results are smoother and more robust to missing points and occlusions with our deformable transformers.
 
\subsection{3D Object Detection}

\begin{table}[]

\begin{center}
\centering
\caption{Spatial transformers improve object detection performance. We report car detection AP(\%) on KITTI validation set. Adding spatial transformers leads to 2\% performance gain.}
\vspace{-1.3em}
\label{tab:detection}
\resizebox{0.5\textwidth}{!}{
\begin{tabular}{>{\columncolor[gray]{0.95}}c|ccc|ccc}
\hline
\rowcolor{LightCyan}
\hline
                      & \multicolumn{3}{c|}{birds' eye} & \multicolumn{3}{c}{3D} \\ \hline
                      \rowcolor{LightCyan2}
                      & Easy          & Medium          & Hard         & Easy       & Medium      & Hard      \\ \hline
VoxelNet\cite{voxelnet}    & 77.3 &        59.6       &     51.6      &     43.8         &    32.6        &             27.9         \\ \hline
VoxelNet + fixed graph  &     84.3    &       67.2     &     59.0   &   45.7    &    34.5     &   32.4    \\ \hline
VoxelNet + deformable &     \textbf{85.3}     &      \textbf{69.1}       &   \textbf{60.9}       &  \textbf{46.1}       &  \textbf{35.9}       &  \textbf{34.0}        \\ \hline

\end{tabular}
}
\vspace{-2em}
\end{center}

\end{table}

We benchmark on KITTI 3D object detection \cite{3dkitti}.  It contains 7,481 training images / point clouds and 7,518 test images / point clouds, covering three categories: Car, Pedestrian, and Cyclist. For each class,
detection outcomes are evaluated based on three difficulty
levels: easy, moderate, and hard,  according to the object size, occlusion state and truncation
level.  We follow the evaluation protocol in VoxelNet \cite{voxelnet} and report car detection results on the validation set.

\noindent \textbf{Network Architectures.}. Shown in Fig.\ref{fig:detection_net}, the network first partitions raw 3D points into voxels. We add deformable spatial transformers; points in each voxel are represented with point features. There are two deformable feature learning layers, each layer having $2$ sub-graphs with $16$-dimensional outputs. Note that the voxel partition is based on the input point coordinates.  As in VoxelNet, the point features in each voxel are fed to $2$ voxel feature encoding layers with channel $32$ and $128$ to get sparse 4D tensors representing the space. The middle convolutional layers process 4D tensors to further aggregate spatial contexts. Finally a Region Proposal Network (RPN) generates the 3D detection.

We report the performance of 3 networks: \begin{inparaenum}[\bfseries (1)] \item VoxelNet baseline \cite{voxelnet}; \item the fixed graph baseline, where we used the original point cloud location to learn the point feature at the place of spatial transformer blocks; \item deformable spatial transformer networks as discussed above. \end{inparaenum}

\noindent \textbf{Results and Analyses.} Table \ref{tab:detection}  reports car detection results on KITTI validation set.\footnote{The authors did not provide code.  We use the implementation by \cite{voxelnetrepo} and obtain lower performance than the original paper.}   Compared with baseline, having a point feature learning module improves the performance by 7.3\% and 2.8\% for birds' eye view and 3D detection performance on average, respectively. The deformable module further improves 8.9\% and 3.9\% respectively over VoxelNet.  

\subsection{Ablation Studies}
We conduct ablation studies to understand how many spatial transformers may be sufficient to achieve satisfactory performance. 
We also study transformations of point coordinates and features of deformable spatial transformers.  The influences of updating transformation matrices and transformers at different layers are investigated.

\label{sec:ablation}
\begin{table}[]

\caption{ Performance of different number of deformable transformation modules. Metric is average mIOU (\%). }
\vspace{-1.3em}
\label{tab:ablation}
\begin{center}

\resizebox{0.45\textwidth}{!}{
\begin{tabular}{>{\columncolor[gray]{0.95}}c|c|c|c|c}
\hline
\rowcolor{LightCyan}

                  & fixed graph & 1 graph & 2 graphs & 4 graphs \\ \hline
$f^{(t)}_i$ = 32 & 84.2   & 84.9   & 85.2   & 85.3   \\ \hline
$f^{(t)}_i k^{(t)}_i$ = 64 & 84.2 & 85.3 & 85.2 &  83.5   \\ \hline

\end{tabular}
}
\end{center}
{\textit{In the first row, the output feature of each sub-graph is of dim. 32, while the number of subgraphs changes; the second row limits  the multiplication of number of sub-graphs and sub-feature dim. to be 64.}}
\vspace{-1em}
\end{table}

\noindent \textbf{The Number of Transformers}.  Table \ref{tab:ablation} shows that for the fixed sub-feature dimension, the more graphs in each layer, the higher the performance. With the fixed complexity, (i.e., the product of the number of sub-graphs and the sub-feature dimension fixed at $64$), the best performance is achieved at  $k=1, f=64$ and $k=2, f=32$ .

\noindent \textbf{Two Components in Deformable transformers}.
A deformable spatial transformer has two components (Equation \ref{eq:deform2}): affine transformation on point coordinates, $AP$, and three-dimensional projection of high-dimensional feature, $CF$. Fig.\ref{fig:deform_ablation} shows that both affine and feature only spatial transformers also improve performance, but the combination of both leads to the largest gain.

\iffalse
\begin{figure}[t]
  \centering
  \includegraphics[width=0.5\textwidth]{figures/numgraph.pdf}
  \caption{Performance of different graphs of deformable spatial transformer.}
  \label{fig:lnumgraph}
\end{figure}
\fi

\begin{figure}
\centering
       \includegraphics[width=0.45\textwidth]{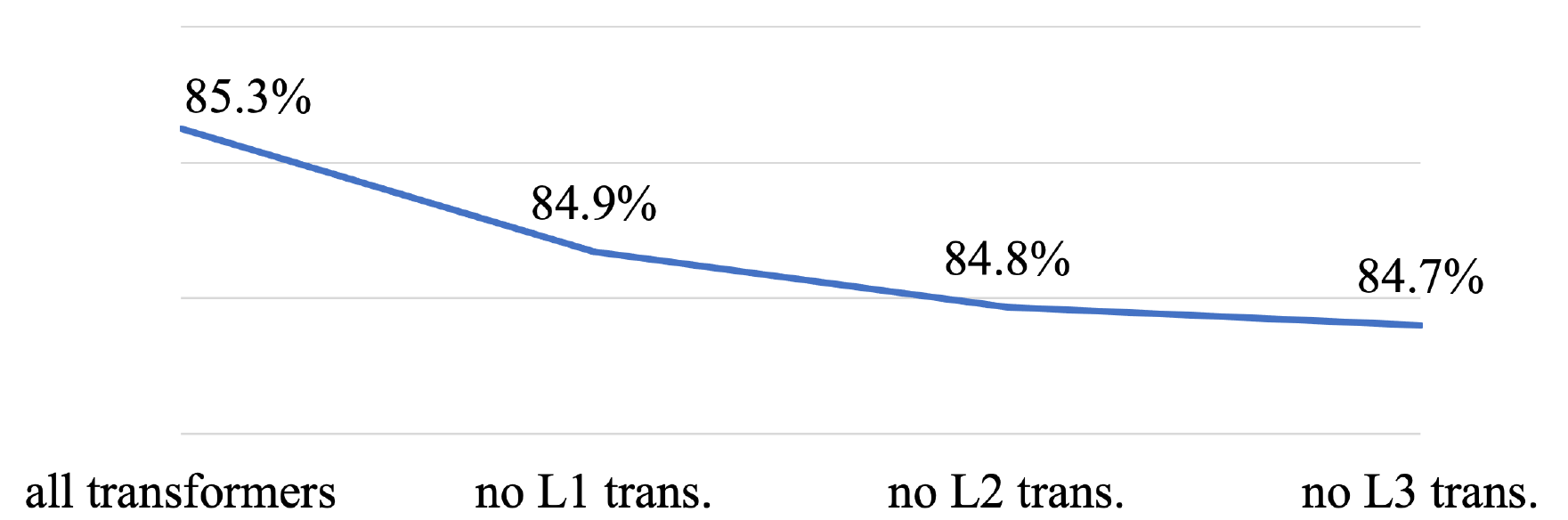}%
   \caption{ \textcolor{black}{Part segmentation performance (average mIOU) of deformable transformers at different layers. When applying all transformers at three layers, the performance is highest. Removing transformers at different layer lead to performance drop. Removing transformers at layer 3 gives the most performance drop.}}
   \vspace{-1em}
   \label{fig:remove}
   \end{figure}

\noindent \textcolor{black}{\textbf{Updating Transformation Matrices}. The transformation matrices are updated in an end-to-end fashion with the ultimate goal of increasing the task performance.  It is of interest to understand if updating transformation matrix boosts the performance.  Specifically, we randomly initialize transformation matrices of deformable spatial transformers and keep them not updated during training.  The performance is 0.5\% better than fixed graphs, indicating that adding more transformation graphs at different layers helps; however, it is 0.6\% worse than updating transformation matrices, indicating learning to update transformation matrices in an end-to-end fashion is helpful.
}

\noindent \textcolor{black}{\textbf{Transformers at Different Layers}. We start with all deformable transformers effective at three layers, and remove transformers at one layer a time.
Fig.\ref{fig:remove} shows that for part segmentation, the performance is best with all transformers, whereas removing transformers at layer 3 gives the largest performance drop, suggesting that transformers at every layer help and those  at the last layer are most important.}

\subsection{Time and Space Complexity}
With spatial transformers, the model size changes little and the inference takes slightly more time (Table \ref{tab:complexity}). 
%Table \ref{tab:complexity} shows that with the same model size and almost the same test time, the significant performance gain can be achieved. 
Note that  for fair comparisons, we increase the number of channels in the fixed graph baseline model for  all the experiments.
Even without increasing the number of parameters of baselines (not shown in Table \ref{tab:complexity}), adding spatial transformers only increases the number of parameters by 0.1\%, as the number of parameters of spatial transformers (only transformation matrices) is very small.
\begin{table}[htb]
\caption{ Model size and test time on ShapeNet part segmentation. Spatial transformers  slightly increase the inference time.}
\vspace{-1.3em}
\label{tab:complexity}
\begin{center}
\resizebox{0.5\textwidth}{!}{
\begin{tabular}{>{\columncolor[gray]{0.95}}c|c|c||c|c}
\hline
\rowcolor{LightCyan}                     & \multicolumn{2}{c}{Sampling-based} & \multicolumn{2}{c}{Point-based}\\ \hline
\rowcolor{LightCyan2}                     & \cite{su2018splatnet}  &  \cite{su2018splatnet} + transformer & \textcolor{black}{ \cite{DGCNN} (fixed)}  &  \textcolor{black}{\cite{DGCNN} + transformer} \\ \hline
\# Params.      & 2,738K   & 2,738K     & \textcolor{black}{2,174K} &   \textcolor{black}{2,174K}                 \\ \hline
Inference time (s/shape) & 0.352     & 0.379      & \textcolor{black}{0.291} &  \textcolor{black}{0.315}                   \\ \hline
\end{tabular}
}
\vspace{-1em}
\end{center}

\end{table}

\subsection{Visualization and Analysis}

We visualize the  change in local neighborhoods when applying spatial transformers. We also visualize the transformed 3D points globally and locally.

\begin{figure}[!t]
  \centering
  \includegraphics[width=0.5\textwidth]{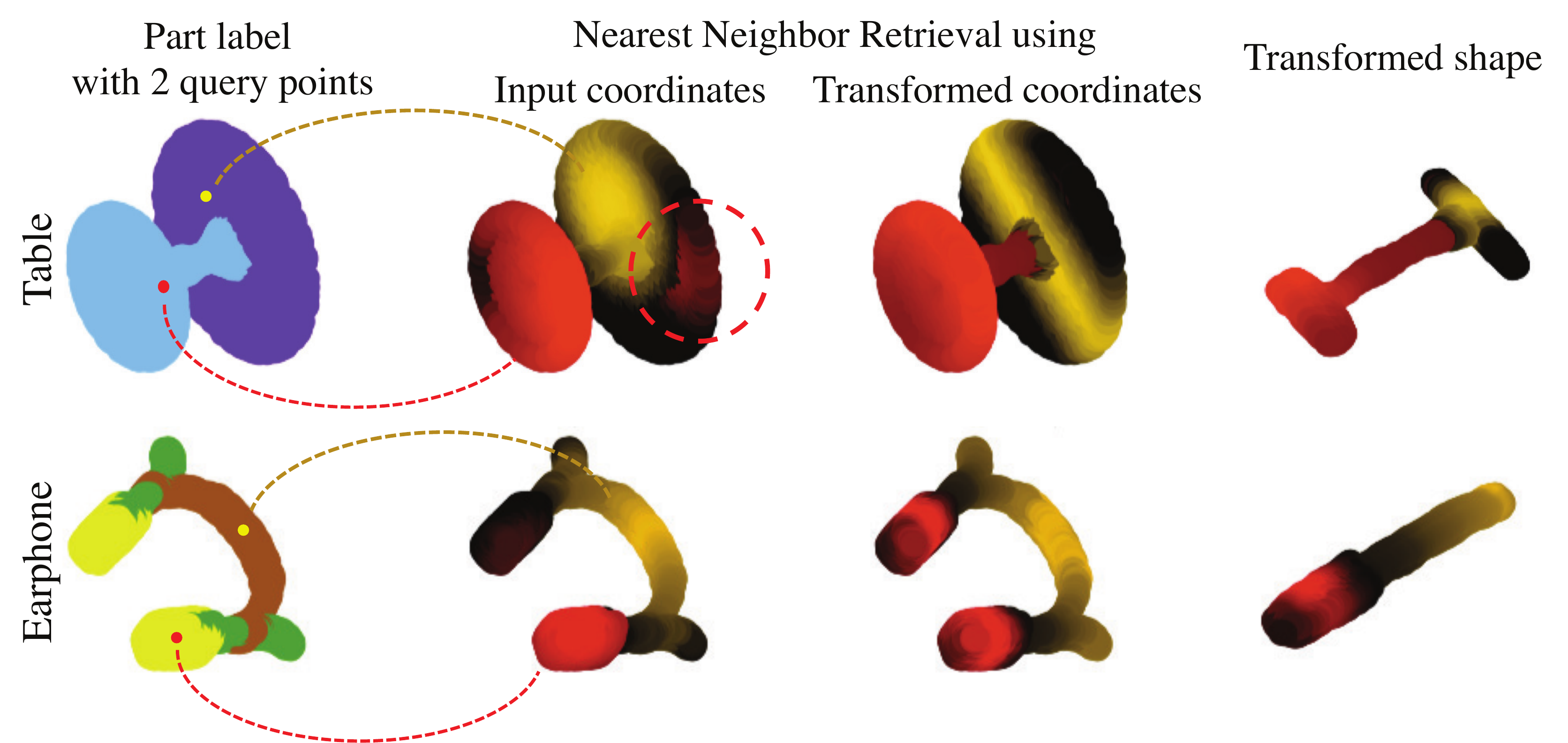}
  \caption{ Local neighborhoods of two query points (red and yellow) using (transformed) 3D coordinates with nearest neighbor retrieval. 
  Neighborhoods of transformed point clouds makes semantic information extraction more efficient: the neighborhood inside the dashed circle adapts to table base part. \href{https://drive.google.com/file/d/1-Y5xp1hFAau6jBryPdD9upm8xR7XTWg8/view?usp=sharing}{View rotating version}  for better visualization. }
  \label{fig:NN}
  \vspace{-1em}
\end{figure}

\noindent \textbf{Dynamic Neighborhood Visualization.}  To illustrate how our spatial transformers learn diverse neighborhoods for 3D shapes, we show the nearest neighbors of two query points and use corresponding colors to indicate corresponding neighborhoods.   \begin{inparaenum}[\bfseries (1)] \item  Fig.\ref{fig:NN} shows that neighborhoods retrieved from deformed shapes encode additional semantic information, compared to neighborhoods from 3D coordinates. \item Fig.\ref{fig:NN2} shows that for \href{https://streamable.com/j2src}{table} and \href{https://streamable.com/rma8m}{earphone}, different graphs enable the network to learn from diverse neighborhoods without incurring additional computational cost. \end{inparaenum}

\begin{figure}[!ht]
  \centering
  \includegraphics[width=0.48\textwidth]{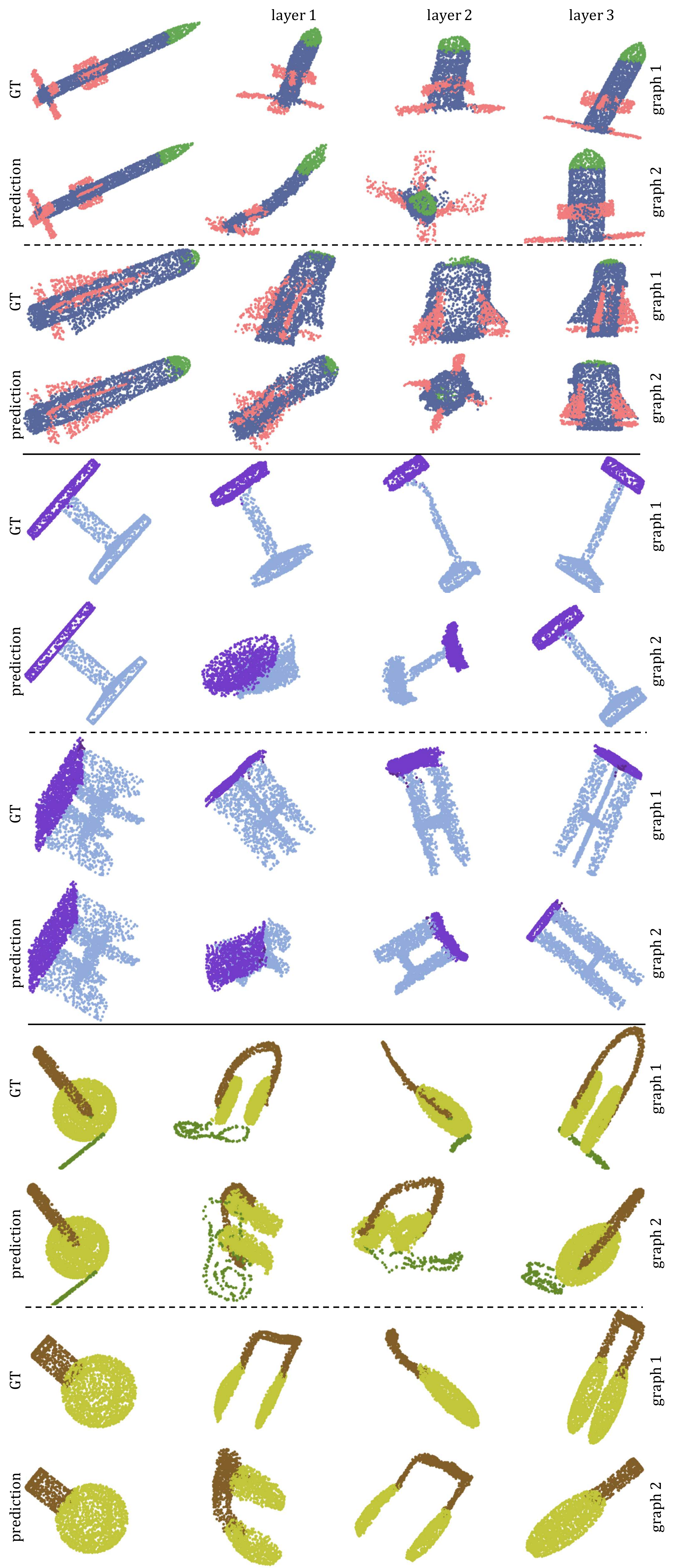}
  \caption{ Examples of learned deformable transformations in ShapeNet part segmentation. 3D shapes include rocket, table and earphone (from up to bottom).  
Every two rows depict an instance with learned transformations.
  We observe that each transformation at certain layer aligns input 3D shape with similar semantic geometric transformation, e.g., graph 2 at layer 2 in rocket examples captures rocket wings. Graph 2 at layer 1 in table examples captures table surfaces.}
    \vspace{-1em}

  \label{fig:part_deform}
\end{figure}
\noindent \textbf{Global Visualization of Deformable Transformations}. Fig.\ref{fig:part_deform} depicts some examples of learned deformable transformations in ShapeNet part segmentation.  Each graph at a certain layer aligns the input 3D shape with similar semantic geometric transformations.  For example, regardless of the shape of the rocket, graph 2 at layer 2 always captures the rocket wing information.

\begin{figure}[!ht]
  \centering
  \includegraphics[width=0.5\textwidth]{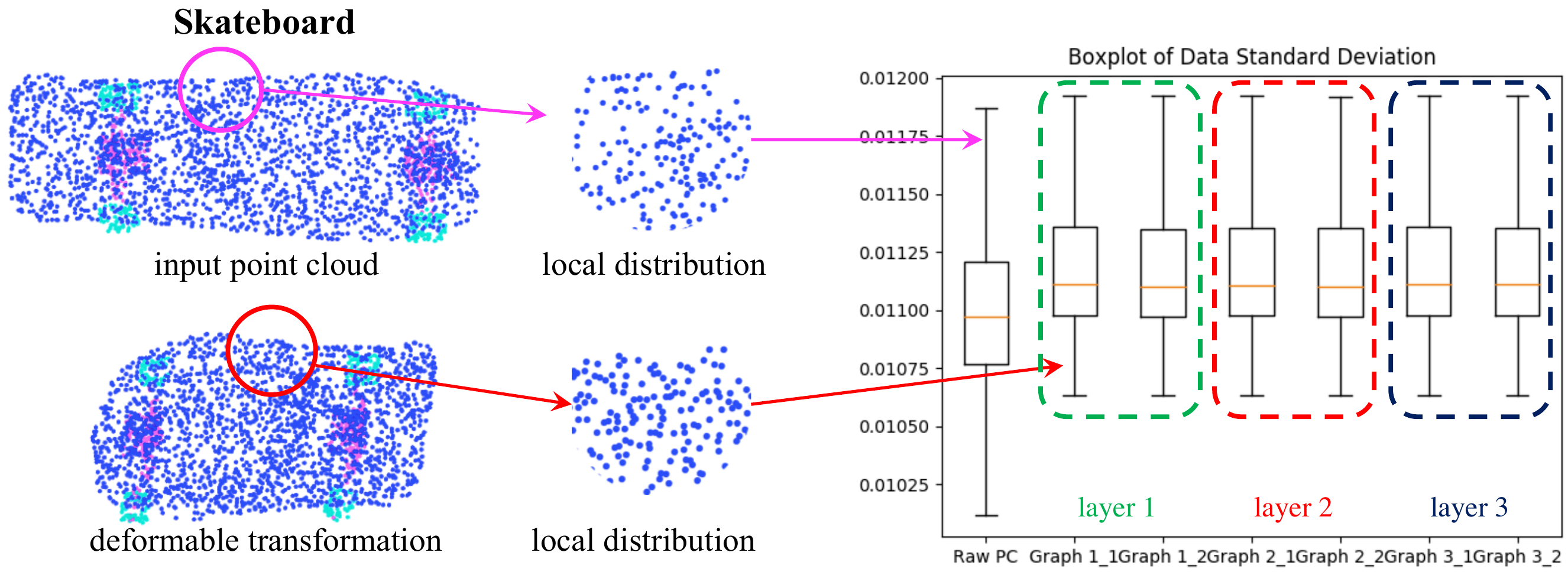}
  \caption{ \textcolor{black}{ 
  Spatial transformers improve the point cloud processing efficiency by improving local distributions of points.
  We show local distributions of a point cloud without and with transformers. The standard deviation of the transformed point cloud is smaller, enhancing the local neighborhood grouping (e.g. when using $k$-NN for affinity matrices, more balanced point distributions make feature learning in each neighborhood suffer less variations and outliers) and feature learning efficiency. }}
  \label{fig:local_view}
    \vspace{-1em}

\end{figure}

\noindent \textbf{Local Distributions after Deformable Transformations}. 3D Points often do not have balanced sampling, which makes point convolution challenging, as the $k$-NN graph does not accurately represents the exact neighborhood and 3D structure information. Our deformable spatial transformer  gives every point flexibility and finds better neighborhoods.  

We wonder if transformers make the point cloud closer to balanced sampling.  We normalize the point coordinates for fair comparisons.  Fig.\ref{fig:local_view} visualizes the local distribution around a sample point on skateboard:  After deformable transformation, the points are moved to a more uniform distribution.  We analyze the standard deviation of raw and transformed \textcolor{black}{point cloud coordinates} in  ShapeNet data.  The standard deviation of point coordinates decreases 50.2\% over all categories after spatial transformations, indicating a more balanced distribution of transformed points. 

We check if the point coordinates are statistically different before and after the application of transformers.  We perform t-test on the original and transformed point clouds. The t-score is 7.15 over all categories with p-value smaller than 1e-9. The transformed point cloud distribution is thus statistically different from the input point cloud distribution.

%% file: sections/5concl.tex
\section{Conclusion}
We propose novel spatial transformers for 3D point clouds that can be easily added onto to existing point cloud processing networks. They can dynamically alter local point neighborhoods for better feature learning.

We study one linear (affine) transformer and two non-linear (projective and deformable) transformers. 
We benchmark them on point-based \cite{DGCNN, li2018pointcnn} and sampling-based \cite{su2018splatnet} point cloud networks and on three
large-scale 3D point cloud processing tasks (part segmentation, semantic segmentation and object detection). Our spatial transformers outperform the fix graph counterpart for state-of-the-art methods.

There are some limitations of our spatial transformers.
First, there are not many constraints on deformable spatial transformers to capture the geometry of the 3D point clouds. 
More complex non-linear spatial transformers may further improve the performance.
On the other hand, spatial transformers learn global transformations of 3D point clouds for altering local neighborhoods.  It is unclear if combining both global and local transformations  \cite{dai2017deformable,thomas2019kpconv} would further improve the learning capacity and task performance.

{ {\bf Acknowledgements.} This research was supported, in part, by Berkeley Deep Drive and DARPA.
The authors thank Utkarsh Singhal and Daniel Zeng for proofreading, and anonymous reviewers for their insightful comments.}
\vspace{-1em}

%% file: biography.tex
\vspace{-4em}

\begin{IEEEbiography}
    [{\includegraphics[width=1in,height=1.25in,clip,keepaspectratio]{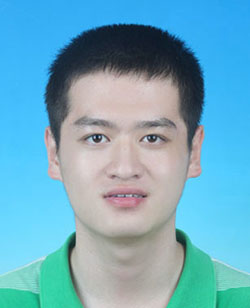}}]{Jiayun Wang} is currently pursing the Ph.D. degree in Vision Science from University of California, Berkeley. He is also a graduate student researcher at the International Computer Science Institute. He received his B.E. degree in Electronic Engineering from Xi'an Jiaotong University, China. He is interested in self-supervised methods for understanding and modeling 3D vision.

\end{IEEEbiography}
\vspace{-3em}
\begin{IEEEbiography}
    [{\includegraphics[width=1in,height=1.25in,trim={0 0 .25in 0},clip,keepaspectratio]{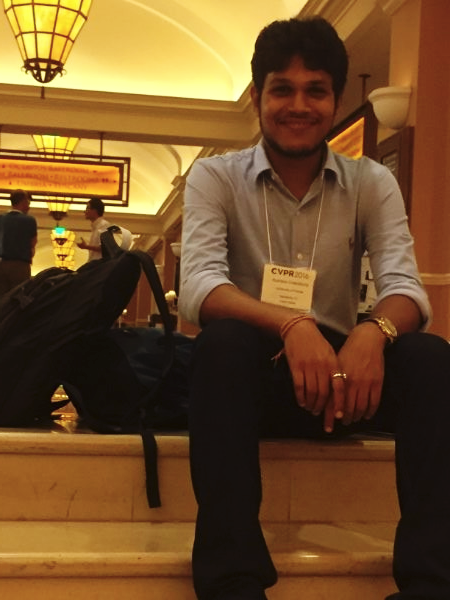}}]{Rudrasis Chakraborty} received his doctorate from the University of Florida, where he studied computer and information science and engineering. He is a member of computer vision group at International Computer Science Institute. He is also a member of Berkeley Deep Drive at UC Berkeley. His research interest includes statistics, differential geometry and deep learning.
\end{IEEEbiography}
\vspace{-3em}
\begin{IEEEbiography}[{\includegraphics[width=1in,height=1.25in,clip,keepaspectratio]{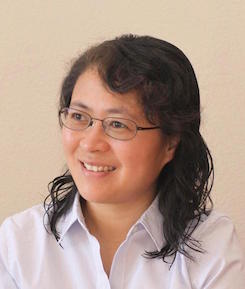}}]{Stella X. Yu}
received her Ph.D. from Carnegie Mellon University, where she studied robotics at the Robotics Institute and vision science at the Center for the Neural Basis of Cognition. She continued her computer vision research as a postdoctoral fellow at UC Berkeley, and then as a Clare Booth Luce Professor at Boston College, during which she received an NSF CAREER award.  Dr. Yu is currently the Director of Vision Group at the International Computer Science Institute (ICSI), a Senior Fellow at the Berkeley Institute for Data Science, and on the faculty of Computer Science, Vision Science, Cognitive and Brain Sciences at UC Berkeley.  She is also affiliated faculty with the Department of Computer and Information Science at the University of Pennsylvania.  Dr. Yu is interested not only in understanding visual perception from multiple perspectives, but also in using computer vision and machine learning to automate and exceed human expertise in practical applications.
\end{IEEEbiography}